%% file: ms.tex
\documentclass[10pt,twocolumn,letterpaper]{article}

\usepackage{etex} 
\usepackage{iccv}
\usepackage{times}
\usepackage{epsfig}
\usepackage{graphicx}
\usepackage{amsmath}
\usepackage{amssymb}

\usepackage{alex}
\usepackage{booktabs} 
\usepackage[caption=false]{subfig} 
\usepackage{algorithm} 
\usepackage{algpseudocode} 
\usepackage{rotating}
\usepackage{enumitem} 
\usepackage{cite} 
\usepackage[font={small}]{caption}

\usepackage{pgfplots}
\pgfplotsset{compat=newest}
\usepackage{tango}
\usepackage{tikz}

\usepgfplotslibrary{colorbrewer}
\definecolor{Dark2-8-1}{RGB}{27,158,119}
\definecolor{Dark2-8-A}{RGB}{27,158,119}
\definecolor{Dark2-8-2}{RGB}{217,95,2}
\definecolor{Dark2-8-B}{RGB}{217,95,2}
\definecolor{Dark2-8-3}{RGB}{117,112,179}
\definecolor{Dark2-8-C}{RGB}{117,112,179}
\definecolor{Dark2-8-4}{RGB}{231,41,138}
\definecolor{Dark2-8-D}{RGB}{231,41,138}
\definecolor{Dark2-8-5}{RGB}{102,166,30}
\definecolor{Dark2-8-E}{RGB}{102,166,30}
\definecolor{Dark2-8-6}{RGB}{230,171,2}
\definecolor{Dark2-8-F}{RGB}{230,171,2}
\definecolor{Dark2-8-7}{RGB}{166,118,29}
\definecolor{Dark2-8-G}{RGB}{166,118,29}
\definecolor{Dark2-8-8}{RGB}{102,102,102}
\definecolor{Dark2-8-H}{RGB}{102,102,102}

\pgfplotsset{compat = 1.3,
         legend style={font=\scriptsize},
         legend cell align={left},
         legend style={cells={align=left}, draw=black!20},
         grid=both,
         grid style={dotted},
         tick style={draw=none},
         enlarge x limits=false,
         enlarge y limits=false,
         axis line style={draw=black!100},}
\pgfplotsset{ every non boxed x axis/.append style={x axis line style=-},
    every non boxed y axis/.append style={y axis line style=-}}

\newlist{inlinelist-roman}{enumerate*}{1}
\setlist*[inlinelist-roman,1]{%
  label=(\roman*),
}
\newlist{inlinelist-alph}{enumerate*}{1}
\setlist*[inlinelist-alph,1]{%
  label=\alph*),
}

\newcommand{\tablestyle}[2]{\setlength{\tabcolsep}{#1}\renewcommand{\arraystretch}{#2}\centering\footnotesize}
\makeatletter\renewcommand\paragraph{\@startsection{paragraph}{4}{\z@}
  {.5em \@plus1ex \@minus.2ex}{-.5em}{\normalfont\normalsize\bfseries}}\makeatother

\newcommand{\KGthree}{\textcolor{black}} 
\newcommand{\KG}{\textcolor{black}}
\newcommand{\KGnew}{\textcolor{black}} 
\newcommand{\KGsupp}{\textcolor{black}} 
\newcommand{\KH}{\textcolor{black}} 
\newcommand{\KHtwo}{\textcolor{black}} 
\newcommand{\KHarxiv}{\textcolor{black}} 
\newcommand{\cc}{\textcolor{black}} 
\newcommand{\cywu}{\textcolor{black}} 
\mathchardef\mhyphen="2D
\graphicspath{{../../images/}}

\usepackage[pagebackref=true,breaklinks=true,letterpaper=true,colorlinks=false,bookmarks=false,hidelinks=true]{hyperref}

\iccvfinalcopy 

\def\iccvPaperID{4311} 

\ificcvfinal\fi
\begin{document}

\title{\vspace{-2.5mm}\KG{Fashion++}: Minimal Edits for Outfit Improvement}

\newcommand\blfootnote[1]{%
  \begingroup
  \renewcommand\thefootnote{}\footnote{#1}%
  \addtocounter{footnote}{-1}%
  \endgroup
}

\author{%
Wei-Lin Hsiao$^{1,4}$ \quad Isay Katsman$^{*2,4}$ \quad Chao-Yuan Wu$^{*1,4}$ \quad Devi Parikh$^{3,4}$ \quad Kristen Grauman$^{1,4}$
\vspace{.5em} \\
$^1$UT Austin  \quad $^2$Cornell Tech \quad $^3$Georgia Tech \quad $^4$ Facebook AI Research
\vspace{-.5em}
}

\maketitle

\begin{abstract}
\blfootnote{$^{*}$ Authors contributed equally.}
Given an outfit, what small changes would most improve its fashionability?  
This question presents an intriguing new vision challenge.
We introduce \emph{Fashion++}, an approach that proposes minimal adjustments to a full-body clothing outfit that will have maximal impact on its fashionability.
Our model consists of a deep image generation neural network that learns to synthesize clothing conditioned on learned per-garment encodings.  The latent encodings are explicitly factorized according to shape and texture, thereby allowing direct edits for both fit/presentation and color/patterns/material, respectively.
We show how to bootstrap Web photos to automatically train a fashionability model, and develop an activation maximization-style approach to transform the input image into its more fashionable self.
The edits suggested range from swapping in a new garment to tweaking its color, how it is worn (e.g., rolling up sleeves), or its fit (e.g., making pants baggier).
Experiments demonstrate that Fashion++ provides successful edits, both according to automated metrics and human opinion.
Project page is at \url{http://vision.cs.utexas.edu/projects/FashionPlus}.
\end{abstract}
\vspace{-1mm}
\input{intro_v2}
\input{related}
\input{approach_v2}
\input{experiments}
\input{conclusion}

{\small
\bibliographystyle{ieee_fullname}
\bibliography{egbib,strings,cvpr2019-refs,cvpr2018-refs}
}

\clearpage
\input{supp}

\end{document}

%% file: intro_v2.tex
\section{Introduction}

\emph{``Before you leave the house, look in the mirror and take one thing off."} -- Coco Chanel
\vspace*{0.05in}

The elegant Coco Chanel's famous words advocate for making small changes with large impact on fashionability.  Whether removing an accessory, selecting a blouse with a higher neckline, tucking in a shirt, or swapping to pants a shade darker, often small adjustments can make an existing outfit noticeably more stylish.  This strategy has practical value for consumers and designers alike.  For everyday consumers, recommendations for how to edit an outfit would allow them to tweak their look to be more polished, rather than start from scratch or buy an entirely new wardrobe.  For fashion designers, envisioning novel enhancements to familiar looks could inspire new garment creations.

\begin{figure}[t]
    \center
    \vspace{-2mm}
    \includegraphics[width=\linewidth]{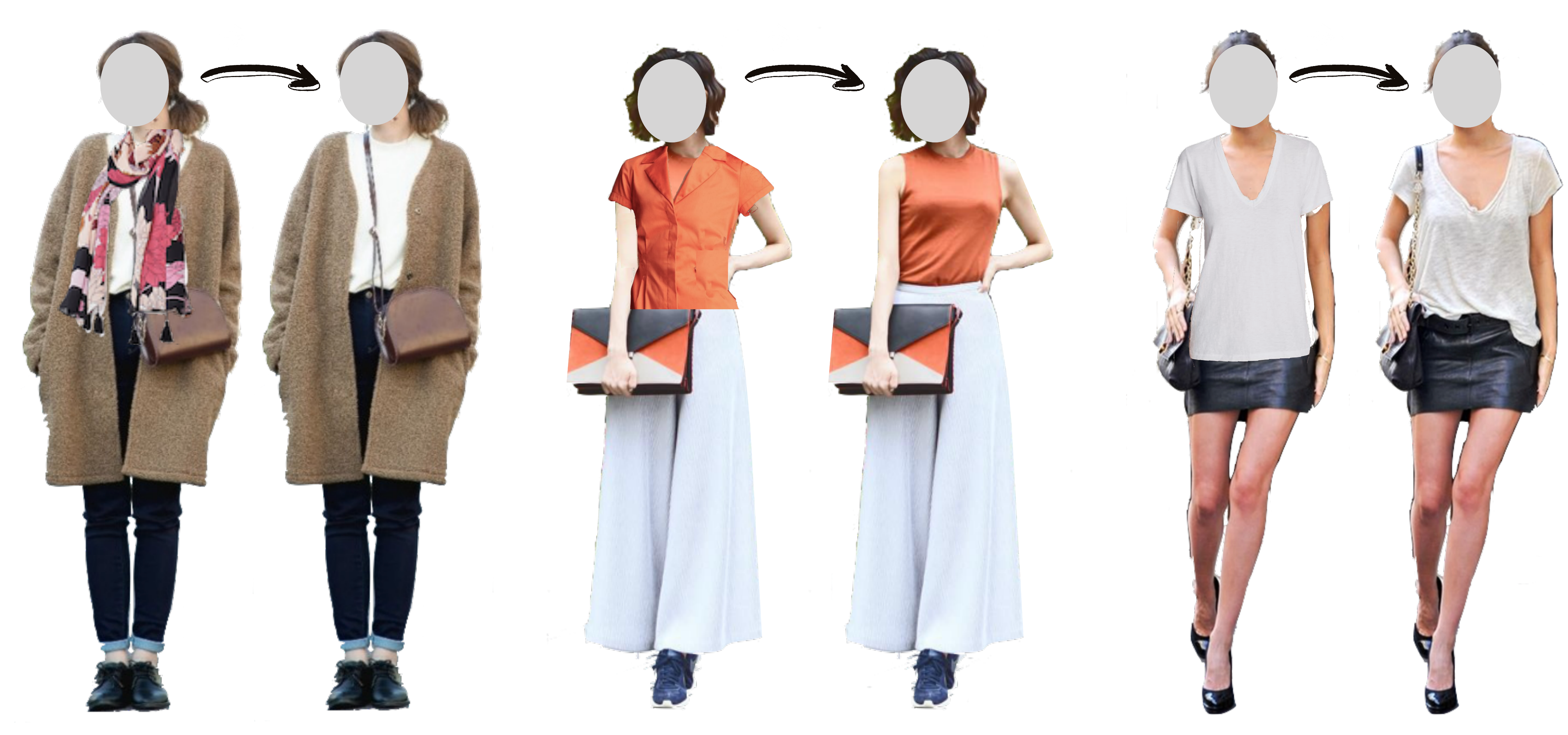}
    \vspace{-7mm}
    \caption{Minimal outfit edits \KGnew{suggest} minor changes to an existing outfit in order to improve its fashionability.  For example, changes might entail (left) removing an accessory; (middle) changing to a blouse with higher neckline; (right) tucking in a shirt.}
  \label{fig:concept}
  \vspace{-3mm}
\end{figure}

Motivated by these observations, we introduce a new computer vision challenge: \emph{minimal edits for outfit improvement}.  
To minimally edit an outfit, an algorithm must propose alterations to the garments/accessories that are slight, yet visibly improve the overall fashionability.   
A ``minimal" edit \KGnew{need} not strictly \KGnew{minimize} the amount of change; \KGnew{rather, it \emph{incrementally adjusts} an outfit} as opposed to starting from scratch.  It can be recommendations on which garment to put on, take off, or swap out, or even how to wear the same garment in a better way.
See Figure~\ref{fig:concept}. 
 
This goal presents several technical challenges.  First, there is the question of training.  A natural supervised approach might curate pairs of images showing better and worse versions of each outfit to teach the system the difference; however, such data is not only very costly to procure, it also becomes out of date as trends evolve.  
Secondly, even with such ideal pairs of images, the model needs to distinguish very subtle differences between positives and negatives (sometimes just \KGnew{a small fraction of} pixels as in \figref{concept}), \cc{which is difficult for an image-based model.} \KGnew{It must reason about the parts (garments, accessories) within the original outfit and how their synergy changes with any candidate tweak.}
Finally, the notion of \emph{minimal} edits implies that adjustments may be sub-garment level, and the inherent properties of the person wearing the clothes---\eg, their pose, body shape---should not be altered.

Limited prior work explores how to recommend a garment for an unfinished outfit~\cite{bilstm,capsule-wardrobe,VasilevaECCV18FashionCompatibility,mcauley-style} (\eg, the fill-in-the-blank task).
Not only \KGthree{is their goal} different from ours, but they focus on clean per-garment catalog photos, and their recommendations \KGnew{are restricted to \emph{retrieved garments} from a dataset.}  However, \KGnew{we observe that} in the fashion domain, the problem demands going beyond seeking an \KGnew{existing} garment to add---\KGnew{to also inferring} which garments are \KGnew{detrimental} 
and should be taken off, and how to adjust the presentation \KGnew{and details} of each garment (\eg, cuff the jeans above the ankle) within a complete outfit to \KGthree{improve its style.}

We introduce a novel image generation approach called Fashion++ to address the above challenges.
The main idea is an activation maximization~\cite{nguyen2016synthesizeprefer} method that operates on localized \KHarxiv{encodings} from a deep image generation 
network. Given an original outfit, we map its composing pieces (\eg, bag, blouse, boots) to their respective \KHarxiv{codes}.  
Then we use a discriminative fashionability model as an editing module to gradually update the \KHarxiv{encoding(s)} in the direction that maximizes the outfit's score, thereby improving its style.
The update trajectory offers a spectrum of edits, starting from the \KGthree{least changed and moving towards} the most fashionable, from which users can choose a preferred end point.
We show how to bootstrap Web photos of fashionable outfits, together with automatically created ``negative" alterations, to train the fashionability model.~\footnote{Fashionability refers to the stylishness of an outfit, the extent to which it agrees with current trends.  As we will see in \secref{trainingdata}, our model defines fashionability by popular clothing choices people wear in Web photos, which can evolve naturally over time with changing trends.}
\KGthree{To account for both the pattern/colors and shape/fit of the garments, 
we factorize each garment's encoding to texture and shape components, allowing the editing module to control where and what to change (\eg, tweaking a shirt's color while keeping its cut vs.~changing the neckline or tucking it in).}

After optimizing the edit, our approach provides its output in two formats: 1) retrieved garment(s) from an inventory
that would best achieve its recommendations 
and 2) a rendering of the same person in the newly adjusted look, generated from the edited outfit's \KHarxiv{encodings}.
\KGthree{Both outputs aim to provide \emph{actionable} advice for small but high-impact changes for an existing outfit.}

\begin{figure*}[t]
  \begin{center}
    \includegraphics[width=\linewidth]{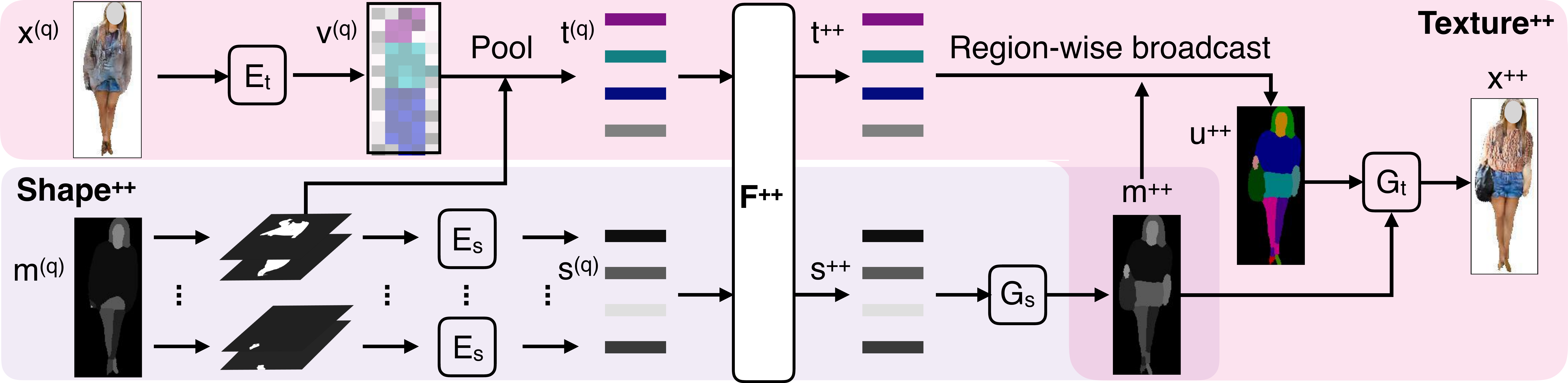}
     \end{center}
    \vspace*{-5mm}
\caption{Overview of our Fashion++ framework. We first obtain latent features from texture and shape encoders $E_t$ and $E_s$. Our editing module $F^{++}$ operates on the latent texture feature $\tb$ and shape feature $\ssb$. After an edit, the shape generator $G_s$ first decodes the updated shape feature $\ssb^{++}$ back to a 2D segmentation mask $\mb^{++}$, and then we use it to region-wise broadcast the updated texture feature $\tb^{++}$ into a 2D feature map $\mathbf{u}^{++}$.  This feature map and the updated segmentation mask are passed to the texture generator $G_t$ to generate the final updated outfit $\xb^{++}$. See Supp. for architecture details.}
  \label{fig:overview}
   \vspace{-4mm}
\end{figure*}

We validate our approach using the Chictopia dataset~\cite{ATR} and, through both automated metrics and user studies, demonstrate \cywu{that} it can successfully generate minimal outfit edits, better than several baselines.   Fashion++ offers a unique new tool for data-driven fashion advice and design---a novel image generation pipeline relevant for a real-world application.

%% file: related.tex
\section{Related Work}
\paragraph{Recognition for fashion.}
Most prior fashion work addresses recognition problems, like 
matching street-to-shop~\cite{street-to-shop2012, where-to-buy-iccv2015, runway-realway, getting-the-look2013}, searching for products interactively~\cite{whittlesearch,dialog-retrieval,zhao-memory-augmented}, and recognizing garments~\cite{deepfashion}.

\paragraph{Fashion image synthesis.}
Synthesis methods explore ways to map specified garments to new poses or people.  This includes 
generating a clothed person conditioned on a product image~\cite{han2018viton, CP-VITON, garmentrecovery2016} (and vice versa~\cite{yoo2016domaintransfer}), or conditioned on textual descriptions (\eg, ``a woman dressed in sleeveless white clothes'')~\cite{fashiongen, zhu2017prada}, as well as methods for swapping clothes between people~\cite{swapnet2018, appperancetransfer2018} or synthesizing a clothed person in unseen poses~\cite{lassner2017generative, Zhao2018multifromsingle, PG2, deformablegans2018, balakrishnan2018synthesizing, chan2018everybodydance, pumarola2018unsupervised}.
Whereas these problems \KG{render} people in a target garment or body pose,
we use image synthesis as a communication tool to make suggestions to minimally edit outfits.

\vspace*{-0.05in}\paragraph{Image manipulation, translation, and style transfer}
are also popular ways to edit images.
There is a large base of literature for generating realistic images conditioned on semantic label maps~\cite{pix2pix2017, CycleGAN2017, BicycleGAN2017, zhu2016manipulation, pix2pixHD2018}, edge maps~\cite{xian2017texturegan, sangkloy2016scribbler}, or 3D models~\cite{liu2017manipulation3d, wu2016latent3d}, using generative adversarial networks (GANs)~\cite{goodfellow2014GAN}.
Related ideas are explored in interactive image search, where users specify visual attributes to alter in their query~\cite{whittlesearch,zhao-memory-augmented,dialog-retrieval}.
Style transfer methods~\cite{Gatys_2015_texture, Gatys_2016_style, gatys_2017_control, huang2017arbitrary} offer another way to edit images that turn photographs into artwork. 
Unlike previous work that conditions on segment maps, maps are \emph{generated} in our case; as a result,
we enable sub-object shape changes that alter regions' footprints, \cc{which generalizes fashion image synthesis.}
Most importantly, all these works aim to edit images according to \emph{human specified input}, whereas we aim to \emph{automatically} suggest where and how to edit to \emph{improve} the input.

\vspace*{-0.05in}
\paragraph{Compatibility and fashionability.}
Fashionability refers to the popularity \KG{or stylishness} of clothing items, while compatibility refers to how well-coordinated individual garments are.
Prior work recommends garments retrieved from a database that go well together~\cite{neurostyle,bilstm,mcauley-compatibility,mcauley-dyadic,iwata,capsule-wardrobe,weilin-iccv2017,VasilevaECCV18FashionCompatibility,craft-ambrish}, or even garments generated from GANs~\cite{shih2017compatibility}. Some also recommend interchangeable items~\cite{bilstm,VasilevaECCV18FashionCompatibility,mcauley-style} that are equally compatible, or forecast future fashion trends~\cite{ziad-iccv2017}. 
\KGthree{We address a new and different problem:} instead of recommending compatible garments from scratch, \KGthree{our approach} tweaks an existing outfit to make it more compatible/fashionable.
\KGthree{It can suggest removals, revise a garment, optimize fashionability, and identify \emph{where} to edit---none of which is handled by existing methods.}
\KG{Using online ``likes" as a proxy for fashionability, the system in}~\cite{fashionability} suggests\KG{---in words---}garments or scenery a user should change to improve fashionability; however, it conditions on  
\KGthree{meta-data rather than images},
and suggests coarse properties specified in words (\eg, Navy and Bags, Black Casual) \KGthree{that} often dictate changing to an entirely new outfit. 

\vspace*{-0.05in}
\paragraph{Activation maximization.}
Activation maximization~\cite{nguyen2016synthesizeprefer} is a gradient based approach that optimizes an image to highly activate a target neuron in a neural network.  It is widely used for visualizing what a network has learned~\cite{nguyen2015fool, simonyan14deep, mahendran2016visualizing, yosinski2017understanding, wei2015understanding}, and recently to synthesize images~\cite{ppgn2017,2017recommendgenerate}. 
In particular, \cite{2017recommendgenerate} also generates clothing images, but they generate single-garment products rather than full body outfits. 
In addition, they optimize images to match purchase history, not to improve fashionability.

%% file: approach_v2.tex
\section{Approach}
Minimal editing 
suggests changes to
an existing outfit 
such that it remains similar but noticeably more fashionable. 
To address this newly proposed task, there are three desired objectives: (1) training must be scalable in terms of supervision and adaptability to changing trends; 
(2) the model could capture subtle visual differences and the complex synergy between \KGthree{garments that affects} fashionability;
and (3) edits should be localized, doing as little as swapping one garment or modifying its properties, while keeping fashion-irrelevant factors unchanged.

In the following, we first present our image generation framework, which decomposes outfit images into their garment regions and factorizes  shape/fit and texture, in support of the latter two objectives (Sec.~\ref{sec:generator}).  Then we present our training data source and discuss how it facilitates the first two objectives (Sec.~\ref{sec:trainingdata}).  Finally, we introduce our activation maximization-based outfit editing procedure and \KHarxiv{show how it recommends garments} (Sec.~\ref{sec:edit}).  

\subsection{Fashion++ Outfit Generation Framework}\label{sec:generator}
The coordination of all composing pieces defines 
an outfit's look. 
To control which parts (shirt, skirt, pants) and aspects (neckline, sleeve length, color, pattern) to change---and also keep identity and other fashion-irrelevant factors unchanged---we want to explicitly model their spatial locality. 
Furthermore, to perform minimal edits, we need to control pieces' \emph{texture} as well as their \emph{shape}.
Texture often decides 
an outfit's theme (style):  
denim with solid patterns gives more casual looks, while leather with red colors gives more street-style looks.
With the same materials, colors, and patterns of garments, how they are worn (\eg, tucked in or pulled out) and the fit (\eg, skinny vs.~baggy pants) and cut (\eg, a V-neck vs.~turtleneck) of a garment will complement a person's silhouette in different ways.
Accounting for all these factors,
we devise an image generation framework that both gives control over individual pieces (garments, accessories, body parts) and also factorizes shape (fit and cut) from texture (color, patterns, materials).

Our system has the following structure at test time:
it first maps an outfit image $\xb^{(q)}$ and its associated semantic \KGthree{segmentation} map $\mb^{(q)}$ to a texture feature $\tb^{(q)}$ and a shape feature $\ssb^{(q)}$.
Our editing module, 
\KH{$F^{++}$, }
then \KGthree{gradually} updates $\tb^{(q)}$ and $\ssb^{(q)}$ into $\tb^{++}$ and $\ssb^{++}$ to improve fashionability.
Finally, based on $\tb^{++}$ and $\ssb^{++}$, the system generates the output \KGthree{image(s)} of the edited outfit $\xb^{++}$. \figref{overview} overviews our system.
Superscripts $(q)$ and $++$ denote variables before and after editing, respectively. We omit the superscript when clear from context.
\KGthree{We next describe how our system maps an outfit into latent features.}
 
\paragraph{Texture feature.}
\vspace{-2mm}
An input image $\xb\in X \subseteq \RR^{H\times W\times C}$ is a real full-body photo of a clothed person.  It is accompanied by a region map $\mb\in M \subseteq \KG{\ZZ}^{H\times W}$ assigning each pixel to a region for a 
clothing piece or body part.  
\KH{We use $n=18$ unique region labels defined in Chictopia10k~\cite{ATR}:}
face, hair, shirt, pants, dress, hats, etc. 
We first feed $\xb$ into a learned texture encoder $E_t : X \rightarrow V$ that outputs a feature map $\mathbf{v} \in V \subseteq \RR^{W\times H \times d_t}$.
\KH{Let $r_i$ be the region associated with label $i$.
We average pool $\mathbf{v}$ in $r_i$ to obtain the texture feature $\tb_i = \mathcal{F}_{pool}^i(\mathbf{v}, \mb) \in \RR^{d_t}, \forall i$.}
\KHtwo{The whole outfit's texture feature is represented as $\tb := [\tb_0; \dots; \tb_{n-1}] \in \RR^{n\cdot d_t}$.}
See \figref{overview} top left.

\paragraph{Shape feature.}
\vspace{-2mm}
We also develop a shape encoding that 
\KH{allows }
per-region shape control separate from texture control.
\KH{Specifically, we construct a binary segmentation map $\mb_i \in \KH{M_B} \in \{0, 1\}^{H \times W}$ for each region $r_i$, 
and use a shared shape encoder $E_s : \KH{M_B}\rightarrow S$ to encode each $\mb_i$ into a shape feature $\ssb_i \in S \in \RR^{d_s}$. 
\KHtwo{The whole outfit's shape feature is represented as $\ssb := [\ssb_0; \dots; \ssb_{n-1}] \in \RR^{n\cdot d_s}$.}
See \figref{overview} bottom left.}

\paragraph{Image generation.}
To generate an image, we first use a shape generator $G_s$ that takes in whole-body shape feature $\ssb$ and generates an image-sized region map $\widehat{\mb} \in M$.
We then perform region-wise broadcasting, which broadcasts $\tb_i$ to all locations with label $i$ based on $\widehat{\mb}$, and obtain the \emph{texture feature map} $\mathbf{u} = \mathcal{F}_{broad}( \tb, \widehat{\mb} ) \in  \RR^{H\times W \times d_t}$.\footnote{\KHarxiv{Note that $\mathbf{u}$ has uniform features for a region, since it is average-pooled, while $\mathbf{v}$ is not.}} 
Finally, we channel-wise concatenate $\mathbf{u}$ and $\widehat{\mb}$ to construct the input to a texture generator $G_t$, which generates the final outfit image.
This generation process is summarized in \figref{overview} (right). 
\KG{Hence, the generators $G_t$ and $G_s$ learn to reconstruct outfit images conditioned on garment shapes and textures.}

\paragraph{Training.}
\KH{Although jointly training the whole system is possible, we found a decoupled strategy to be effective.}
Our insight is that if we assume a fixed semantic region map, the generation problem
is reduced to an extensively studied image translation problem, and we can benefit from recent advances in this area.
In addition, if we separate the shape encoding and generation from the whole system,
it reduces to an auto-encoder, which is also easy to train.

Specifically, 
for the image translation part (Texture++ in \figref{overview}), we adapt from conditional generative adversarial networks (cGANs) that take in segmentation label maps and associated feature maps to generate photo-realistic images~\cite{BicycleGAN2017,pix2pixHD2018}.
We  combine the texture encoder $E_t$ and texture generator $G_t$ with a discriminator $D$ to formulate a cGAN.
\KHarxiv{An image $\widehat{x}$ is generated by $G_t(\mb,\mathbf{u})$, }
where $\mathbf{u} = \mathcal{F}(E_t(\xb),\mb)$, and $\mathcal{F}$ is the combined operations of $\mathcal{F}_{pool}^i, \forall i$ and $\mathcal{F}_{broad}$.
The discriminator $D$ aims to distinguish real images from generated ones.  $E_t$, $G_t$ and $D$ are learned simultaneously with a minimax adversarial game objective:
\begin{equation}
  {G_t}^\ast\!,{E_t}^\ast\!\!=\!\! \argmin_{G_t,E_t} \max_{D} \Lcal_{\textrm{GAN}}(G_t,\!D,\!E_t) \KH{+\Lcal_{\textrm{FM}}(G_t,\!E_t,\!D)},
\end{equation}
\KG{where $\Lcal_{\textrm{GAN}}$ is defined as:}
\begin{equation}
    \mathbb {E}_{(\mb,\xb)}\rbrr{\log D(\mb,\xb) +\log\rbrr{1-D\rbrr{\mb,G_t\rbr{\mb,\mathbf{u}}}}}
\end{equation}
for all training images $\xb$,
\KH{and $\Lcal_{\textrm{FM}}$ denotes feature matching loss.} 
For the shape deformation part of our model (Shape++ in \figref{overview}), we formulate a shape encoder and generator with a region-wise Variational Autoencoder (VAE)~\cite{vae2013}.
The VAE assumes the data is generated by a directed graphical model 
$p(\mb | \ssb )$ 
and the encoder learns an approximation $q_{\KG{E_s}}(\ssb |\mb )$ to the posterior distribution \KH{$p(\ssb |\mb )$.} The prior over the encoded feature is  set to be Gaussian with zero mean and identity covariance, $p(\ssb )=\Ncal(\mathbf {0,I} )$. 
The objective of our VAE is to minimize the Kullback-Leibler ($\mathrm {KL}$) divergence between $q_{E_s}(\ssb |\mb )$ and $p(\ssb )$, and the $\ell_1$ reconstruction loss:
\begin{equation}
D_{\mathrm {KL} }\rbrr{q_{E_s}(\ssb |\mb ) \Vert  p(\ssb )} + \KH{\mathbb {E}_{\mb} \normbrr{\mb-G_s\rbrr{E_s(\mb)}}_1.}
\end{equation}

Note that simply passing in the 2D region label map \KH{as the shape encoding $\ssb$} would be insufficient for image editing.  The vast search space of all possible masks is too difficult to model, and, during editing, mask alterations could  often yield unrealistic or uninterpretable ``fooling" images~\cite{nguyen2015fool, simonyan14deep}.
In contrast, our VAE design learns the probability distribution of the outfit shapes, and hence can generate unseen shapes corresponding to variants of features from the learned distribution.  This facilitates meaningful shape edits.

Having defined the underlying image generation architecture, we next introduce our 
editing module for revising an input's \KHarxiv{features (encodings)} to improve fashionability.

\subsection{Learning Fashionability from Web Photos}\label{sec:trainingdata}

Our editing \KHarxiv{module} 
(Sec.~\ref{sec:edit}) requires a discriminative model of fashionability, which prompts the question:
how can we train a fashionability classifier for minimal edits?
Perhaps the ideal training set would consist of pairs of images in which each pair shows the same person in slightly different outfits, one of them judged to be more fashionable than the other.  However, such a collection is not only impractical to curate at scale, it would also become out of date as soon as styles evolve.
\KH{An alternative approach is to treat a collection of images from a specific group (\eg, celebrities) as positive exemplars and another group (\eg, everyday pedestrians) as negatives. However, we found such a collection suffers from conflating identity and style, and thus the classifier finds fashion-irrelevant properties discriminative between the two groups.}

\begin{figure}[t]
    \center
    \includegraphics[width=\linewidth]{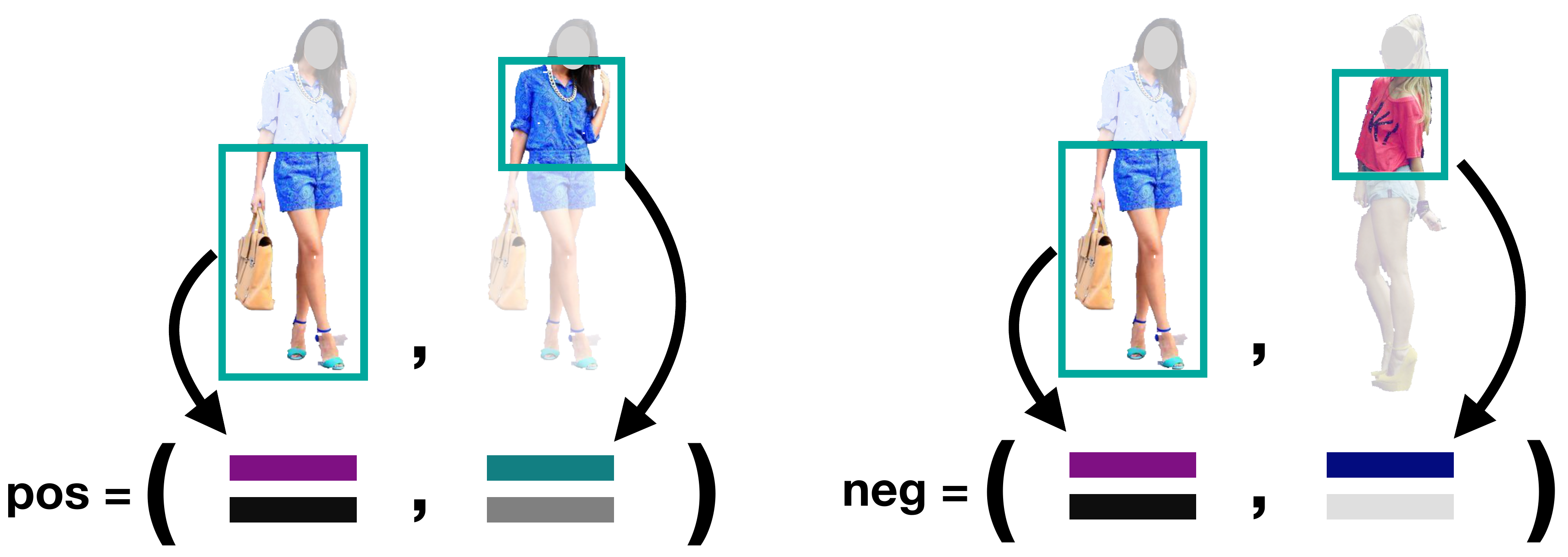}
    \vspace*{-7mm}
    \caption{Forming training examples: \KGthree{A fashionable Web photo is the positive (left). We overwrite some garment's features with those from another \emph{distant} outfit to create the negative (right).  (Here only two of $n$ garment regions are shown for simplicity.)}}
    \label{fig:outfit_manipulation}
     \vspace*{-0.1in}
\end{figure}

Instead, we propose to bootstrap less fashionable photos automatically from Web photos of fashionable outfits.
The main idea is to create 
\KH{``negative"} outfits from fashionista photos.  We start with a Chictopia full-body outfit photo (a ``positive"), select one of its pieces to alter, and  replace it with a piece from a different outfit.  To increase the probability that the replacement piece degrades fashionability, we extract it from an outfit that is most dissimilar to the original one, as measured by Euclidean distance on CNN features.  We implement the garment swap 
\KGthree{by} overwriting the \KHarxiv{encoding} $\zb_i := [\tb_i; \ssb_i]$ for garment $i$ with the target's.  See \figref{outfit_manipulation}. 

\KG{We use this data to train a 3-layer multilayer perceptron (MLP) fashionability classifier $f$.  It is trained to map the \KHarxiv{encoding} $\zb:= [\tb; \ssb]$ for an image $\xb$ to its binary fashionability label $y \in \{0,1\}$.}

The benefit of this training strategy is threefold: First, it makes curating data easy, and also refreshes easily as styles evolve---by downloading new positives. Second, by training the fashionability classifier on these decomposed (to \KGthree{garments}) and factorized (shape vs.~texture) \KHarxiv{encodings}, a simple MLP effectively captures the subtle visual properties and complex garment synergies (see Supp.\ for ablation study). Finally,
we stress that our approach learns from full-body outfit photos being worn by people on the street, as opposed to clean catalog photos of \KGthree{individual} garments~\cite{neurostyle,bilstm,mcauley-compatibility,mcauley-dyadic,shih2017compatibility,VasilevaECCV18FashionCompatibility}.  This has the advantages of allowing us to learn aspects of fit and presentation (\eg, tuck in, roll up) that are absent in catalog data, as well as the chance to capture organic styles based on what outfits people put together in the wild.  

\subsection{Editing an Outfit}\label{sec:edit}

With the \KH{encoders $E_t, E_s$, generators $G_t, G_s$} and \KHarxiv{editing module $F^{++}$} 
in hand, we now explain how our approach performs a minimal edit.
Given test image $\xb^{(q)}$, 
Fashion++ returns its edited \KGthree{version(s):}
\begin{equation}
 \xb^{++} := G\rbr{F^{++}\rbr{E\rbrr{\xb^{(q)}}}},
\end{equation}
where $G$ and $E$ represent the models for both shape and texture.
When an inventory of discrete garments is available, our approach also returns the nearest real garment $g_i^{++}$ 
for region $i$ that could be used to achieve that change, as we will show in results.
\KGthree{Both outputs---the rendered outfit and the nearest real garment---are complementary ways to provide actionable advice to a user.}

\vspace*{-0.03in}

\input{section33}

%% file: section33.tex
\paragraph{Computing an edit.} 
The main steps are: calculating the desired edit, and generating the edited image.
To calculate an edit,
we take an activation maximization approach: we iteratively alter the 
outfit's feature such that it increases the activation of the fashionable label according to $f$.

\begin{figure}
\centering
   \includegraphics[width=.75\linewidth]{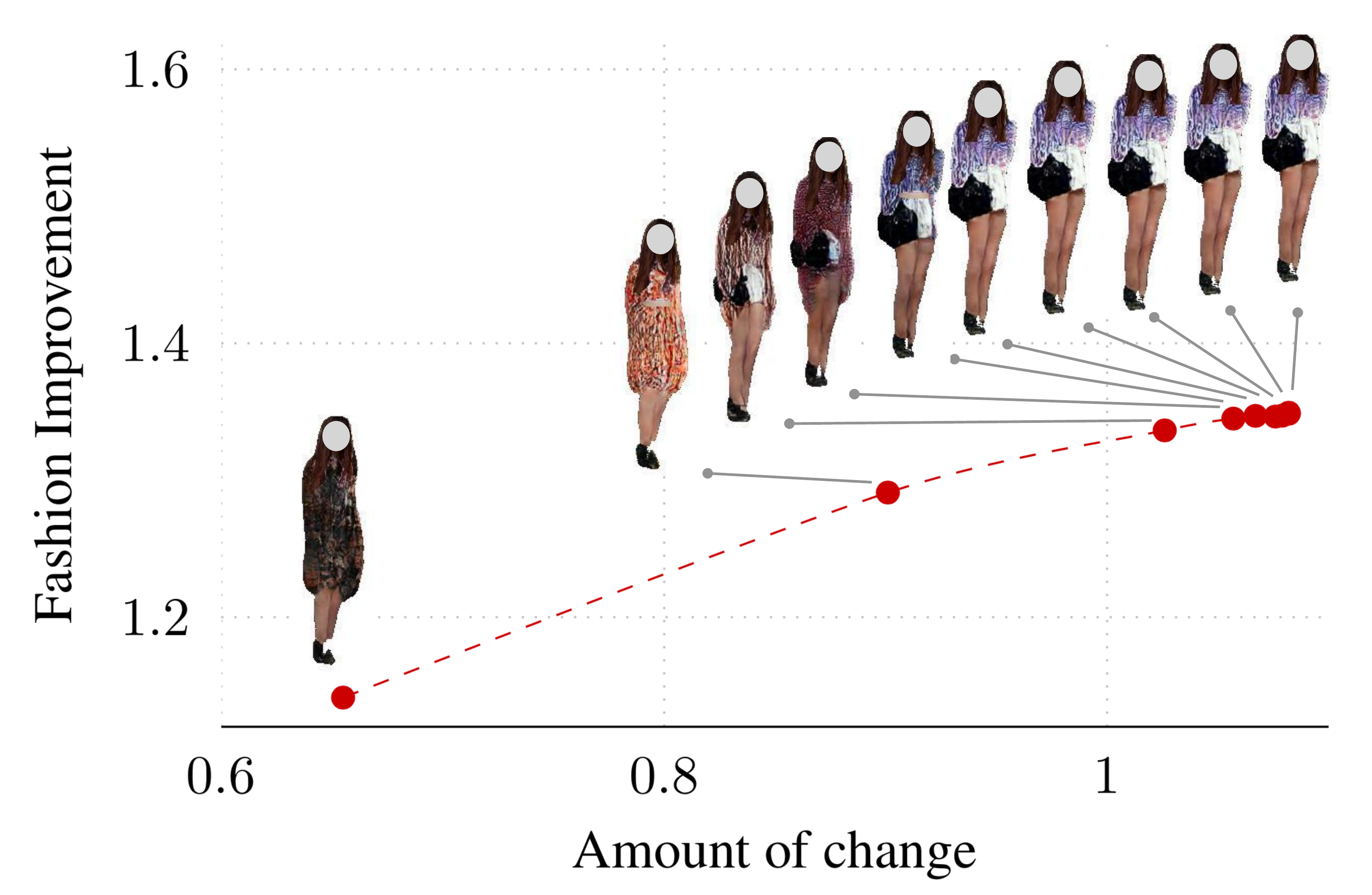}
    \vspace*{-0.15in}
    \caption{As Fashion++ iteratively edits the outfit, the fashionability improves and eventually saturates as the outfit becomes fashionable enough. (Metrics defined in~\secref{auto_quant}.  Dots show average result for \emph{all} test images.)}
    \label{fig:auto_ours_iter}
    \vspace*{-2mm}
\end{figure}

Formally, let $\zb^{(0)} := \{\tb_{0}, \ssb_{0}, \dots, \tb_{n-1}, \ssb_{n-1}\}$
be the set of all features in an outfit, 
and $\tilde{\zb}^{(0)} \subseteq \zb^{(0)}$ be a subset of features corresponding to the \emph{target regions or aspects} that are being edited (\eg, shirt region, shape of skirt, texture of pants). 
We update the outfit's representation as:
\begin{equation}
\tilde{\zb}^{(k+1)} := \tilde{\zb}^{(k)} + \lambda \frac{\partial{p_f\rbrr{y=1| \zb^{(k)} }}}{\partial{\tilde{\zb}^{(k)}}}, k = 0, \dots, K-1 
\label{eq:update}
\end{equation}
where $\tilde{\zb}^{(k)}$ denotes the features after $k$ updates, 
$\zb^{(k)}$ denotes substituting only the target features in $\zb^{(0)}$ with $\tilde{\zb}^{(k)}$ \KH{while keeping other features unchanged},
$p_f(y=1|\zb^{(k)})$ denotes the probability of fashionability according to classifier $f$, and $\lambda$ denotes the update step size. 
Each gradient step in \eqnref{update} yields an incremental adjustment to the input outfit. 
\figref{auto_ours_iter} shows the process of taking $10$ gradient steps with step size $0.1$ (see \secref{exp} for details).
By presenting this spectrum of edits to the user, one may choose a preferred end point 
(i.e., his/her preferred tradeoff in the ``minimality" of change vs.~maximality of fashionability).
Finally, as above, $\zb^{(K)}$ gives the updated $\tb_i^{++}; \ssb_i^{++}, \forall i$.

To further force updates to stay close to the original, one could add a proximity objective, $\normbr{\zb^{(k)}-\zb^{(0)}}$, as in other editing work \cite{liu2017manipulation3d, zhu2016manipulation}. However, balancing this smoothness term with other terms (users' constraints in their cases, fashionability in ours) is tricky (\eg, \cite{liu2017manipulation3d} reports non-convergence).  We found our gradient step approach to be at least as effective to achieve gradual edits.

\vspace*{-0.02in}

\paragraph{Optimizing where to edit.}
A garment for region $i$ is represented as the concatenation of its texture and shape features: $\zb_i^{(0)} := [\tb_i; \ssb_i]$.
Our approach optimizes 
the garment that ought to be edited 
by cycling though all garments to find the one with most impact:
\begin{equation}
  i^* = \argmax_{i=0, \dots, n-1} \normbrr{\frac{\partial{p_f(y=1|\zb^{(0)})}}{\partial{{\zb_i}^{(0)}}}}.
  \label{eq:where_update}
\end{equation}
By instructing the target $\tilde\zb^{(0)}$ to be ${\zb_{i^*}}^{(0)}$, we can simultaneously optimize \emph{where and how to change} an outfit.

\vspace*{-0.01in}

\paragraph{Rendering the edited image.}
Then we generate the Fashion++ image output by conditioning our image generators $G_t, G_s$ on these edits:
\begin{equation}\vspace*{-0.025in}
  \KH{\xb^{++}} = G_t({\mb}^{++}, \mathbf{u}^{++}),
  \vspace*{-0.025in}
\end{equation}
where $\mathbf{u}^{++}$ refers to the broadcasted 
map of the edited texture components 
$\tb^{++}$, and 
${\mb}^{++} = G_s(\ssb^{++}) $
refers to the VAE generated mask for the edited shape components $\ssb^{++}$. The full edit operation is outlined in \figref{overview}.

In this way, our algorithm automatically updates the latent encodings to improve fashionability, then passes its revised code to the image generator to create the appropriate image.  
An edit could affect as few as one or as many as $n$ garments, and 
we can control whether edits are permitted for shape or texture or both.  This is useful, for example, if we wish to insist that the garments look about the same, but be edited to have different tailoring or presentation (e.g., roll up sleeves)---shape changes only.  

\vspace*{-0.01in}
\paragraph{Retrieving a real garment matching the edit.}
Finally, we return the garment(s) $g_i^{++}$ that optimally achieves the edited outfit.  Let $\Ical$ denote an inventory of garments.
The best matching garments to retrieve from 
$\Ical$ 
are:
\begin{equation}
{g_i}^{++} := \argmin_{g_i \in \Ical}{\normbrr{\zb_{g_i}-{\zb_i}^{++}}},
\label{eq:retrieve}
\end{equation}
for $i=0,\dots,n-1$, 
where $\zb_{g_i}$ denotes the garment's feature.
This is obtained by passing the real inventory garment image for $g_i$ to the texture and shape feature encoders $E_t$ and $E_s$, and concatenating their respective results.

%% file: experiments.tex
\section{Experiments}
\label{sec:exp}
We now validate that Fashion++ 
\begin{inlinelist-roman}
  \item makes slight yet noticeable improvements better than baseline methods in both quantitative evaluation (\secref{auto_quant}) and user studies (\secref{human_study});
  \item \cc{effectively communicates to users through image generation (\secref{human_study}); and}
  \item supports all possible edits from swapping, adding, removing garments to adjusting outfit presentations via qualitative examples (\secref{qual_results}).
\end{inlinelist-roman}
\paragraph{Experiment setup.}
We use the Chictopia10k~\cite{ATR} dataset for all experiments. We use $15,930$ images to train the generators, and $12,744$ to train the fashionability classifier. 
We use the procedure described in \secref{trainingdata} to prepare positive and negative examples for training the fashionability classifier.
We evaluate on $3,240$ such unfashionable \KGthree{examples.}
We stress that all test examples are from \emph{real world outfits}, bootstrapped by swapping \emph{features} (not pixels) of pieces from different outfits.  
\KGthree{This allows testing on real data while also having ground truth (see below).}
We use the region maps provided with Chictopia10k for all methods, though automated semantic segmentation could be used. 
Model architectures and training details are in Supp.

\paragraph{Baselines.}
Since our work is the first to consider the minimal edit problem, we develop several baselines for comparison:
\textsc{Similarity-only}, which selects the nearest neighbor garment in the database $\Ical$ (Chictopia10k) to maintain the least amount of change; \textsc{Fashion-only}, which changes to the piece that gives the highest fashionability score as predicted by our  classifier, using the database $\Ical$ as candidates; \textsc{Random sampling}, which changes to a randomly sampled garment.
Since all unfashionable outfits are generated by swapping out a garment, we instruct all methods to update that garment.
We additionally run results where we automatically determine the garment to change, denoted auto-Fashion++.

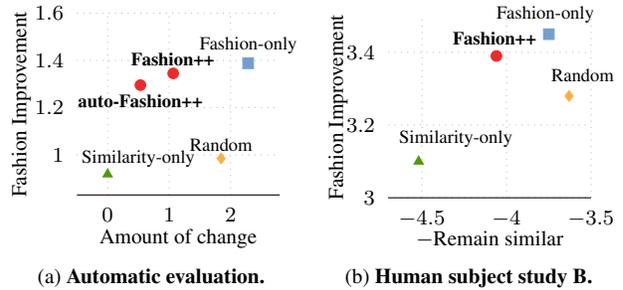
\begin{figure}
  \captionsetup[subfigure]{labelformat=empty}
  \footnotesize
  \subfloat[(a) \textbf{Automatic evaluation.} \label{fig:auto_baselines}] {
      \begin{tikzpicture}
        \begin{axis}[xlabel=Amount of change,
                     ylabel=Fashion Improvement,
                     xlabel shift = -3pt,
                     ylabel shift = -4pt,
                     axis lines=left,
                     xmin=-0.5,
                     ymin=0.83,
                     xmax=2.8,
                     ymax=1.6,
                     width=0.245\textwidth,
                     height=4cm,
                     y axis line style={draw opacity=0},
                    ]
               \addplot[
                       visualization depends on={value \thisrow{name}\as\myvalue},
                   scatter/classes={
                       b_data={mark=square*,skyblue1,text mark=\myvalue},
                       c_data={mark=triangle*,chameleon3,text mark=\myvalue},
                       d_data={mark=diamond*,orange1,text mark=\myvalue},
                       b_name={mark=x,white,text mark=\myvalue},
                       c_name={mark=x,white,text mark=\myvalue},
                       d_name={mark=x,white,text mark=\myvalue}
                       },
                       scatter, only marks,
                       scatter src=explicit symbolic,
                       nodes near coords*={\myvalue},font=\scriptsize,]
                table[x=x,y=y,meta=label]{auto_quant_min_baselines_grounded.txt};
              \addplot[
                       visualization depends on={value \thisrow{name}\as\myvalue},
                   scatter/classes={
                       a1_data={mark=*,scarletred1,text mark=\myvalue},
                       a1_name={mark=*,white,text mark=\myvalue},
                       a2_data={mark=*,scarletred1,text mark=\myvalue},
                       a2_name={mark=*,white,text mark=\myvalue}
                       },
                       scatter, only marks,
                       scatter src=explicit symbolic,
                       nodes near coords*={\textbf{\myvalue}},font=\scriptsize,]
                table[x=x,y=y,meta=label]{auto_quant_min_ours_grounded.txt};
        \end{axis}
      \end{tikzpicture}
    }
  \hfill
  \subfloat[(b) \textbf{Human subject study \KGthree{B}.} \label{fig:new_median}] {
      \begin{tikzpicture}
        \begin{axis}[xlabel=$-$Remain similar,
                     ylabel=Fashion Improvement,
                     xlabel shift = -4pt,
                     ylabel shift = -6pt,
                     axis lines=left,
                     xmin=-4.7,
                     ymin=3.0,
                     xmax=-3.5,
                     ymax=3.5,
                     xtick={-5,-4.5,...,-3},
                     width=0.245\textwidth,
                     height=4cm,
                     y axis line style={draw opacity=0},
                    ]
               \addplot[
                       visualization depends on={value \thisrow{name}\as\myvalue},
                   scatter/classes={
                       b_data={mark=square*,skyblue1,text mark=\myvalue},
                       c_data={mark=triangle*,chameleon3,text mark=\myvalue},
                       d_data={mark=diamond*,orange1,text mark=\myvalue},
                       b_name={mark=x,white,text mark=\myvalue},
                       c_name={mark=x,white,text mark=\myvalue},
                       d_name={mark=x,white,text mark=\myvalue}
                       },
                       scatter, only marks,
                       scatter src=explicit symbolic,
                       nodes near coords*={\myvalue},font=\scriptsize,]
                table[x=x,y=y,meta=label]{protocol_a_baselines.txt};
                 \addplot[
                       visualization depends on={value \thisrow{name}\as\myvalue},
                   scatter/classes={
                       a_data={mark=*,scarletred1,text mark=\myvalue},
                       a_name={mark=*,white,text mark=\myvalue}
                       },
                       scatter, only marks,
                      scatter src=explicit symbolic,
                      nodes near coords*={\textbf{\myvalue}},font=\scriptsize,]
                table[x=x,y=y,meta=label]{protocol_a_ours.txt};
        \end{axis}
      \end{tikzpicture}
     }
     \vspace*{-0.1in}
  \caption{For both automatic (a) and human (b) evaluation, Fashion++ best balances improving fashionability while remaining similar.  In (b), both axes are the raw Likert scale; we negate the x-axis so that its polarity agrees
  \KH{to the left.}}
  \label{fig:auto_quant}
  \vspace*{-4mm}
\end{figure}

\subsection{Quantitative comparison}
\label{sec:auto_quant}

Minimal edits change an outfit by improving its fashionability while not changing it too much. Thus, we evaluate performance simultaneously by \emph{fashionability improvement} and \emph{amount of change}.
We evaluate the former by how much the edit gets closer to the ground-truth (GT) outfit.
Since each unfashionable outfit is generated by swapping to a garment (we will call it original) from another outfit, and the garment before the swap (we will call it GT) is just one possibility for a fashionable outfit, we form a \emph{set of GT garments} per test image, representing the multiple ways to improve it
\KH{(see Supp.\ for details).}
The \emph{fashion improvement} metric is the ratio of the original piece's distance to the GT versus the edited piece's distance to the GT. Values less than one mean no improvement.
The \emph{amount of change} metric scores the edited garment's 
\KH{distance }
to the original garment, normalized by subtracting \textsc{Similarity only}'s number. All distances are Euclidean distance in the generators' \KHarxiv{encoded} space. 
All methods return the garment in the inventory nearest to their predicted \KHarxiv{encoding}.

\figref{auto_baselines} shows the results.\KHarxiv{\footnote{We plot ours with $K=6$ for clarity \KGthree{and since fashionability typically saturates soon after}.  Results for all $K$ values are in \figref{auto_ours_iter} \KGthree{and Sec.~\ref{sec:human_study}.}}}
\textsc{Similarity-only} changes the outfit the least, as expected, but it does not improve fashionability. \textsc{Fashion-only} improves fashionability the most, but also changes the outfit significantly. \textsc{Random} neither improves fashionability nor remains similar. Our Fashion++ improves fashionability nearly as well as the \textsc{Fashion-only} baseline, while remaining as similar to the original outfit as \textsc{similarity-only}. 
\KH{Auto-Fashion++ performs similarly to Fashion++.}  \KGthree{These results support our claim that Fashion++ makes slight yet noticeable improvements.}

\figref{auto_ours_iter} shows that by controlling the amount of change (number of gradient steps) made by Fashion++, one can choose whether to \emph{change less} (\KGthree{while still being more} fashionable than \textsc{similarity-only}) or \emph{improve fashionability more} (\KGthree{while still changing less than \textsc{fashion-only}}).

\subsection{Human perceptual study}
\label{sec:human_study}
\input{section42.tex}

\subsection{Minimal edit examples}
\label{sec:qual_results}
Now we show example outfit edits. We first compare side-by-side with the baselines, and then show variants of Fashion++ to demonstrate its flexibility.
\KGthree{For all examples, we show outfits both before and after editing as reconstructed by our generator.}
\vspace*{-0.05in}
\paragraph{General minimal edits comparing with baselines.}
\figref{general} shows examples of outfit edits by all methods \KGthree{as well as the retrieved nearest garments.}
Both \textsc{fashion-only} (ii) and \textsc{random} (iv) change the outfit a great deal. While \textsc{random} makes outfits less fashionable, \textsc{fashion-only} improves them with more stylish garments. Fashion++ (i) also increases fashionability, and the recommended change bears similarity (in shape and/or texture) to the initial less-fashionable outfit. For example, the bottom two instances in~\figref{general} wear the same shorts with different shirts.  \textsc{fashion-only} recommends changing to the same white blouse with a red floral print for both instances, which looks fashionable but is entirely different from the initial shirts; Fashion++ recommends changing to a striped shirt with a similar color palette for the first one, and changing to a sleeveless shirt with a slight blush for the second.
\textsc{Similarity-only} (iii) indeed looks similar to the initial outfit, but stylishness also remains similar.

\begin{figure}
   \center
    \includegraphics[width=\linewidth]{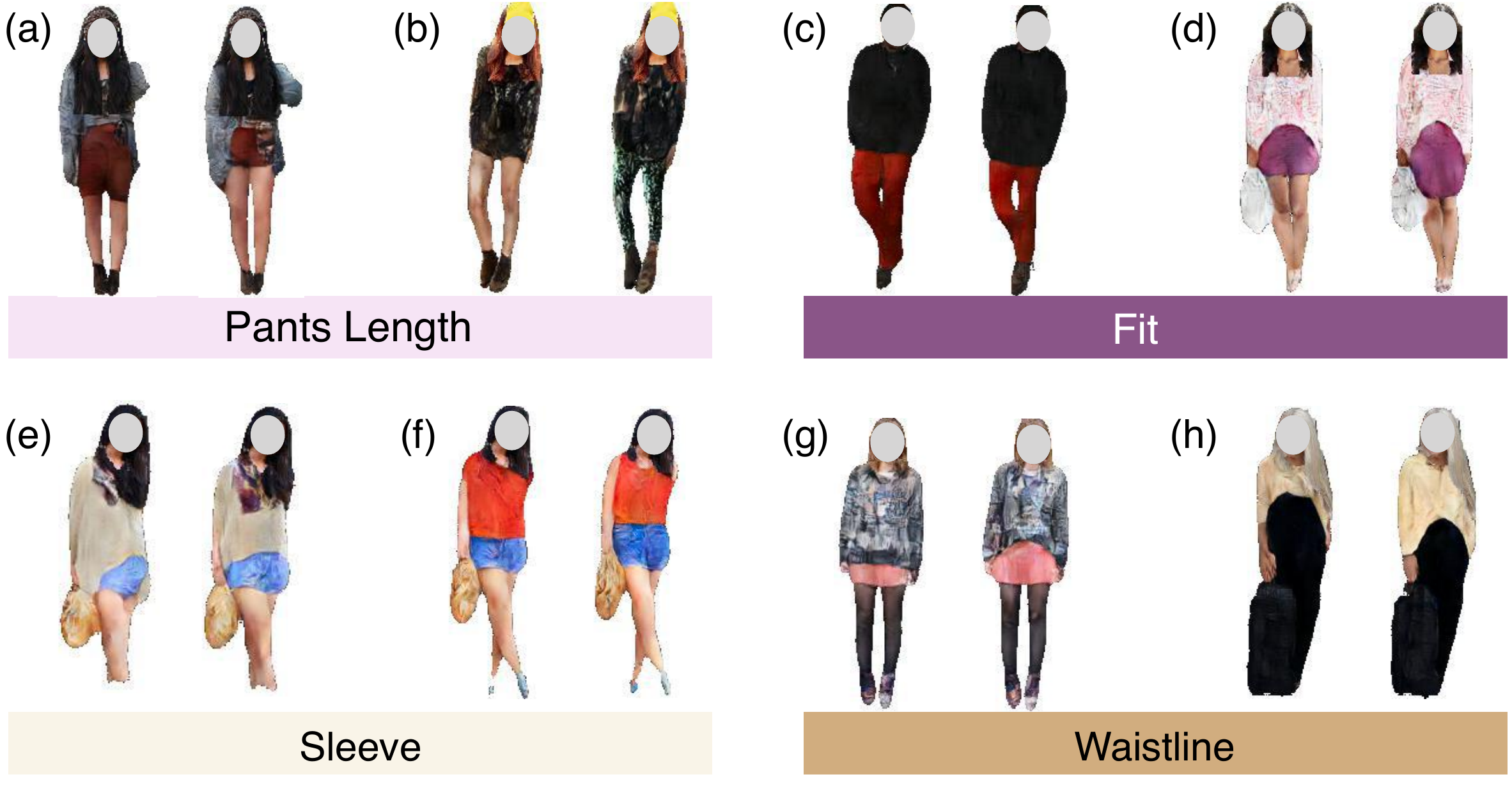}
  \caption{Fashion++ minimal edits on only shape/fit.}
  \label{fig:shape_change}
  \vspace*{-0.1in}
\end{figure}

\vspace*{-0.05in}
\paragraph{Minimal edits changing only shapes.}
\KH{\figref{shape_change} shows examples when we instruct our model to just change the shape }
(cf.~Sec~\ref{sec:edit}). Even with the exact same pieces and person, adjusting the clothing proportions and fit can favorably affect the style.
\figref{shape_change} (a) shows the length of pants changing. Notice how changing where the shorts end on the wearer's legs lengthens them.
(b,c) show changes to the fit of pants/skirt: wearing pieces that fit well emphasizes wearers' figures.
(d) wears the same jacket in a more \KGthree{open} fashion that gives character to the look.
(e,f) roll the sleeves up: slight as it is, it makes an outfit more energetic (e) or dressier (f).
(g,h) adjusts waistlines: every top and bottom combination looks different when tucked tightly (g) or bloused out a little (h), and properly adjusting this for different ensembles gives better shapes and structures.

\begin{figure}
  \center
     \includegraphics[width=\linewidth]{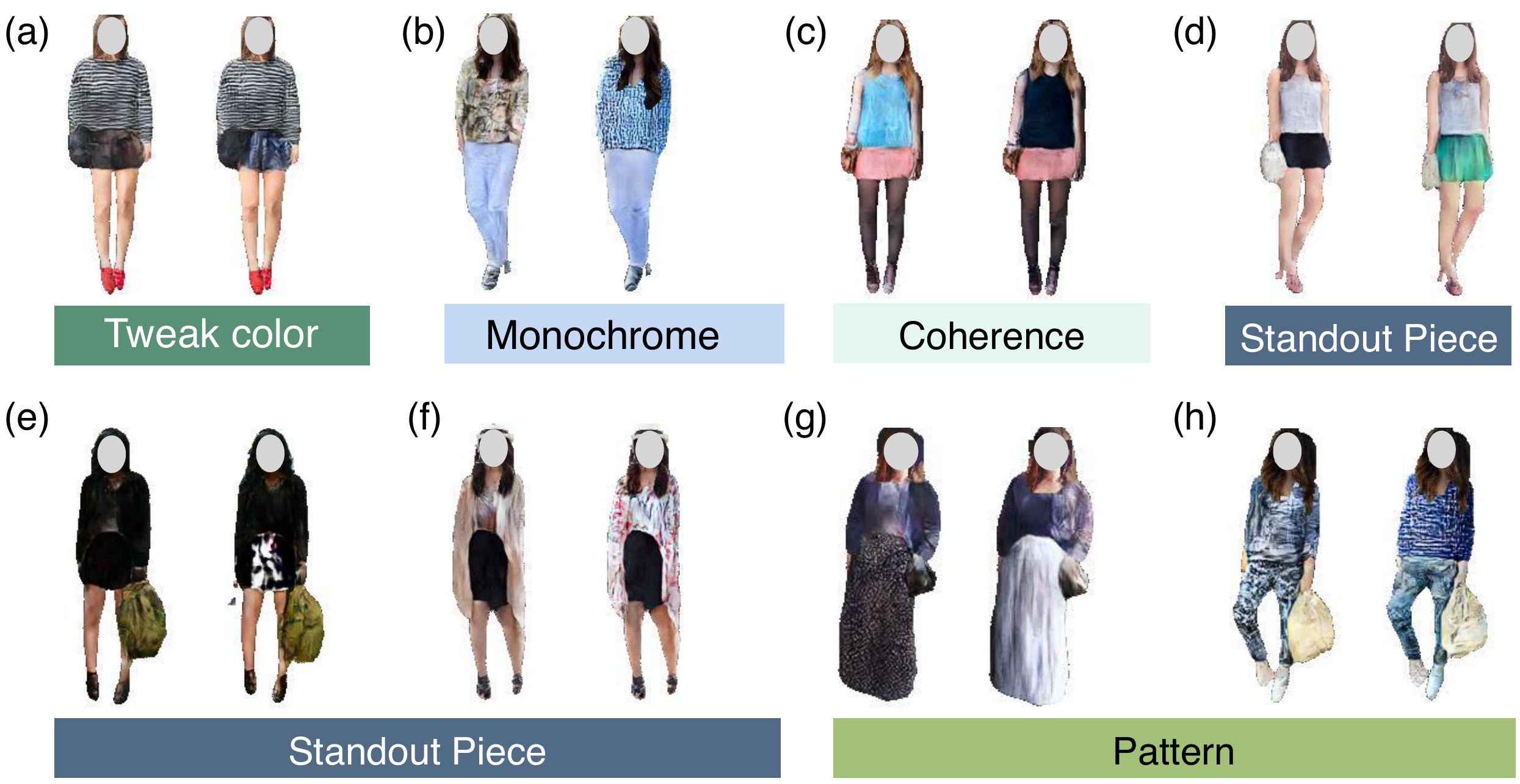}
    \caption{Fashion++ minimal edits on only color/pattern.
    }
    \label{fig:texture_change}
\end{figure}

\paragraph{Minimal edits changing only textures.}
\KH{\figref{texture_change} shows examples when we instruct our model to just change the texture.}
(a) polishes the outfits by changing the bottom a tint lighter. (b) changes the outfit to a monochrome set that lengthens the silhouette.
(c) swaps out the incoherent color. (d)-(f) swap to stand-out pieces
by adding bright colors or patterns that make a statement for the outfits. 
(g)-(h) are changing or removing patterns: notice how even with the same color components, changing their proportions can light up outfits in a drastic way.

\begin{figure}
   \center
    \includegraphics[width=.95\linewidth]{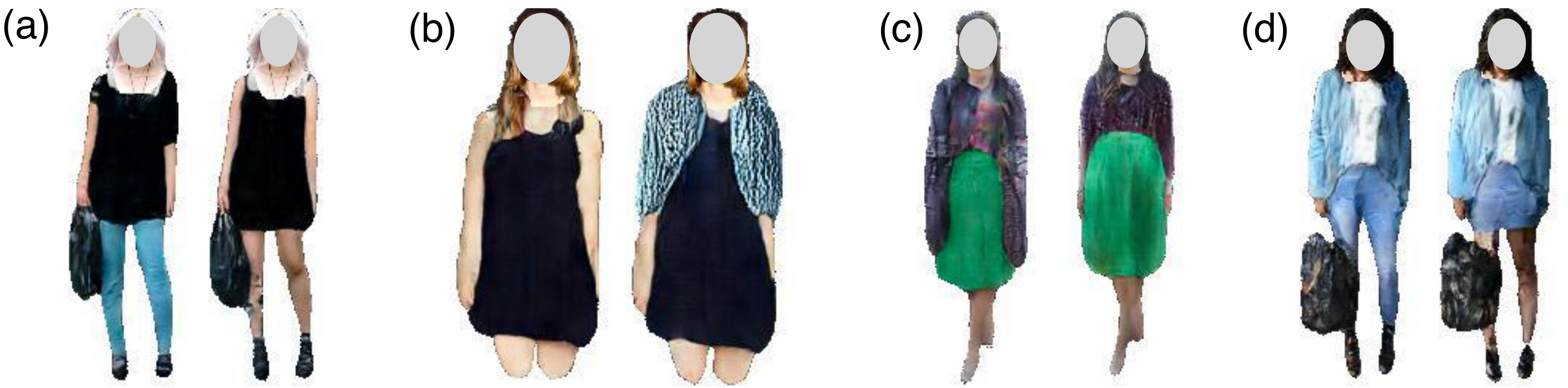}
   \vspace{-3mm}
   \caption{Fashion++ edits that add/remove clothing pieces.}
  \label{fig:add_remove}
\end{figure}

\paragraph{Beyond changing existing pieces.}
Not only can we tweak pieces that are already on outfits, but we can also take off redundant pieces and even put on new pieces.  \figref{add_remove} shows such examples. 
In (a), the girl is wearing a stylish dress, but together with somewhat unnecessary pants.
(b) suggests to add outerwear to the dress for more layers, while (c) takes off the dark outerwear for a lighter, more energetic look.
(d) changes pants to skirt for a better figure of the entire outfit.

\begin{figure}
     \centering
     \subfloat[][]{\includegraphics[scale=.18]{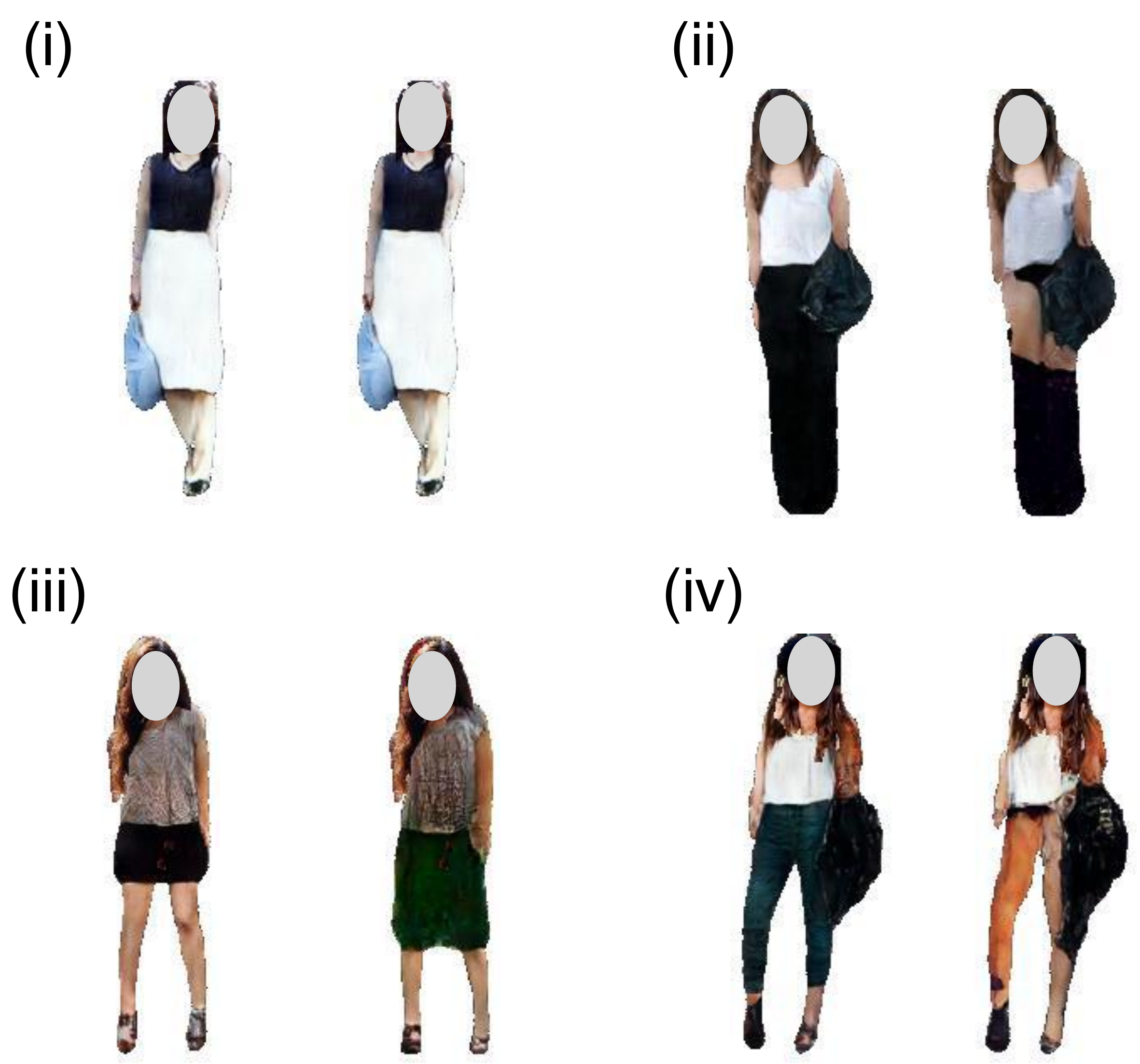}\label{fig:failure}}\hfill
     \subfloat[][]{\includegraphics[scale=.18]{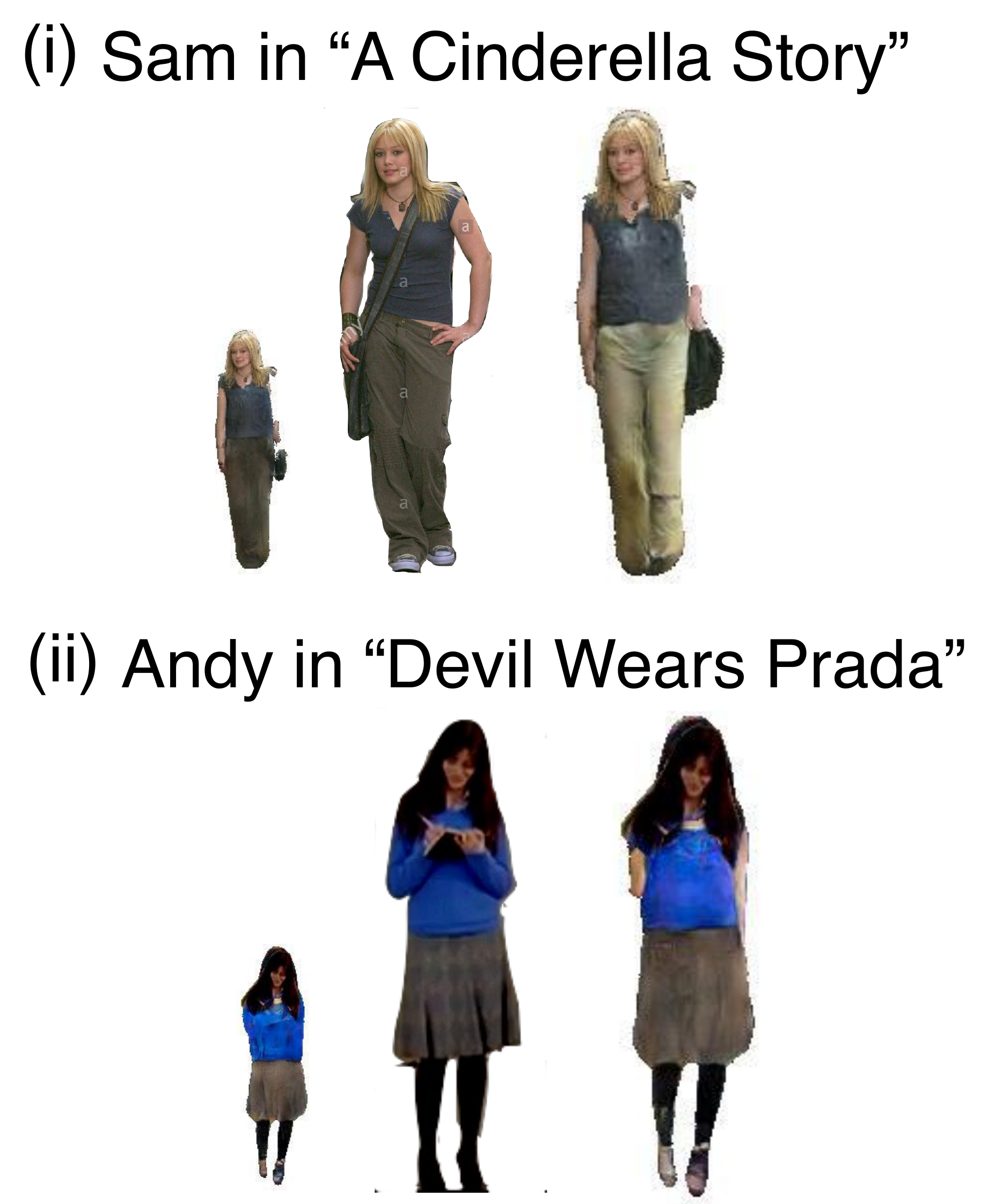}\label{fig:real_edits}}
     \vspace*{-2mm}
     \caption{(a): Some failure cases of Fashion++; (b): Fashion++ on \KGthree{notoriously unfashionable characters.}\vspace*{2mm}}
     \label{fig:failure_and_reals}
     \vspace*{-5mm}
\end{figure}

\paragraph{Failure cases.}
A minimal edit requires good outfit generation models, an accurate fashionability classifier, and robust editing operations. Failure in any of these aspects can result in worse outfit changes. \figref{failure} shows some failure examples as judged by Turkers.

\paragraph{\KGthree{Editing celebrities.}}
\figref{real_edits} shows Fashion++ operating on movie characters known to be unfashionable. 

%% file: section42.tex
Next we ask humans to judge the quality of Fashion++'s edits, how it compares with baselines, and whether they know what actions to take to improve outfits based on the edits.
We perform three human subject test protocols; \KGthree{please see Supp.~for all three user interfaces.}
We randomly sample $100$ unfashionable test outfits and post tasks on Mechanical Turk (MTurk).
Each sample is answered by 7 people, and in total $282$ Turkers answered.

\paragraph{Protocol A.}
Fashion++ can show users a spectrum of edits (\eg, \figref{auto_ours_iter}) from which to choose the desired version. \KGthree{While preference will naturally vary} among users, we are interested in knowing 
\KGthree{to what extent} a \KGthree{given degree of} change is preferred and why.  To this end, we show Turkers an original outfit and edits from $K=1$ to $10$, and ask them to:
\begin{inlinelist-roman}
  \item Select all edits that are more fashionable than the original.
  \item Choose which edit offers the best balance in improving the fashionability without changing too much.
  \item Explain why the option selected in (ii) is best.
\end{inlinelist-roman}

For (i), we found that the more we change an outfit (increasing $K$), the more often human judges think the changed outfit becomes fashionable, with $92\%$ of the changed outfits 
judged as more fashionable when $K=10$.
Furthermore, when we apply Fashion++ to an already fashionable outfit, $84\%$ of the time the human judges find the changed outfit to be similarly or more fashionable, meaning Fashion++ ``does no harm'' in most cases (see Supp.).
For (ii), no specific $K$ dominates. 
The top selected $K=2$ is preferred $18\%$ of the time, and $K=1$ to $6$ are each preferred at least $10\%$ of the time.
This suggests that results for $K\leq6$ are similarly representative, so we use $K=6$ for remaining user studies. 
For (iii), a common reason for a preferred edit is being more \emph{attractive}, \emph{catchy}, or \emph{interesting}.
See Supp.\ for detailed results breaking down $K$ for (i) (ii) and more Turkers' verbal explanations for (iii).

\paragraph{Protocol B.}
Next we ask human judges to compare Fashion++ to the baselines defined above.
We give workers a pair of images at once: 
one is the original outfit and the other is edited by a method (Fashion++ or a baseline). They are asked to express their agreement with two statements on a five point Likert scale: 
\begin{inlinelist-roman}
  \item The changed outfit is more fashionable than the original.
  \item The changed outfit remains similar to the original.
\end{inlinelist-roman}
We do this survey for all methods.
We report the median of the 7 responses for each pair. 

\begin{figure}[t]
   \begin{center}
     \includegraphics[width=\linewidth]{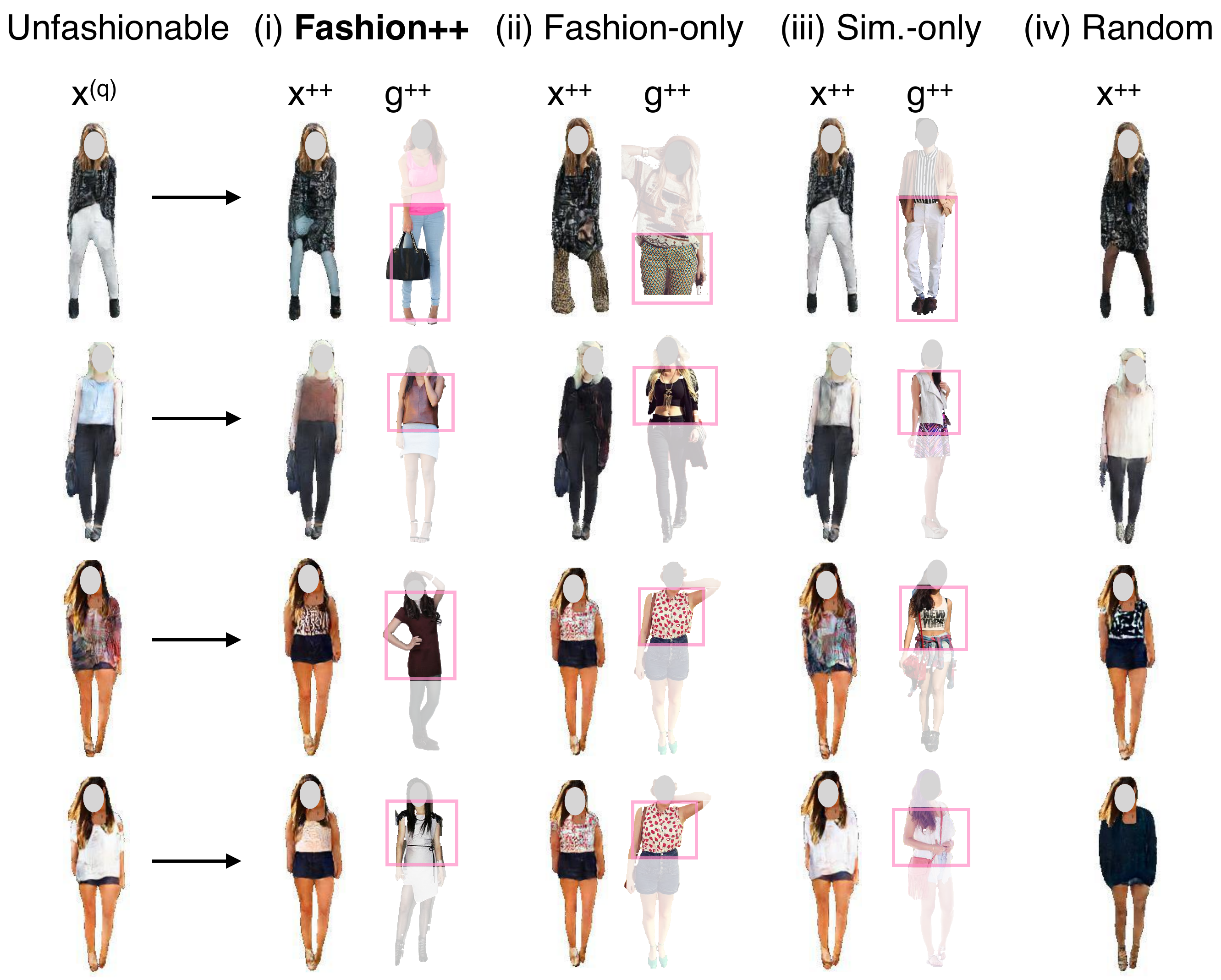}
    \end{center}
    \vspace*{-5mm}
    \caption{Minimal edit comparisons with baselines. Rows are instances, columns are results for methods:
    For all but \textsc{random} (iv), we show both the rendered (left) and retrieved (right) results. Retrieved garments $g_i^{++}$ are in bounding boxes. Best on pdf.}
   \label{fig:general}
\end{figure}

\figref{new_median} shows the result.
It aligns very well with our quantitative evaluation \KGthree{in \figref{auto_baselines}}: \textsc{Fashion-only} is rated as improving fashionability the most,
but it also changes outfits as much as \textsc{random}.
\textsc{similarity-only} is rated as remaining most similar.
Fashion++ changes more than \textsc{similarity-only} but less than all others, while improving fashionability nearly as much as \textsc{fashion-only}. This \KGthree{strongly reinforces} that Fashion++ makes \KGthree{edits that are slight yet improve fashionability.}

\paragraph{Protocol C.}
Finally, it is important that no matter how good the image's exact pixel quality is, humans can get \emph{actionable information} from the suggested edits to improve outfits. We \KHarxiv{thus} ask Turkers how ``actionable'' our edit is on a five point Likert scale, \cc{and to verbally describe the edit.} 
$72\%$ of the time human judges find our images actionable, rating the clarity of the actionable information as $4.16\pm0.41 / 5$. ($4$ for \emph{agree} and $5$ for \emph{strongly agree}). See Supp.~for Turkers' verbal descriptions of our edits.

%% file: conclusion.tex
\section{Conclusions}

We introduced the minimal fashion edit problem.  Minimal edits are motivated by consumers' need to tweak existing wardrobes and designers' desire to use familiar clothing as a springboard for inspiration.  We introduced a novel image generation framework to optimize and display minimal edits yielding more fashionable outfits, accounting for essential technical issues of locality, scalable supervision, and flexible manipulation control.  Our results are quite promising, both in terms of quantitative measures and human judge opinions.
In future work, we plan to broaden the composition of the training source, e.g., using wider social media platforms like Instagram~\cite{snavely-street-style}, bias an edit towards an available inventory, or generate improvements conditioned on an individual's preferred style or occasion.

%% file: supp.tex
 {\LARGE Supplementary Material}
 \vspace{5mm}

\setcounter{section}{0}
\renewcommand\thesection{\Roman{section}}

This supplementary file consists of:
\begin{itemize}
    \item Implementation details of the complete Fashion++ system presented in Section 4 of the main paper
    \item Ablation study on our outfit's representation (referenced in Section 3.2 of the main paper)
    \item Details on shape generation
    \item \KG{More details on} the automatic evaluation metric defined in Section 4.1 of the main paper
    \item \KHarxiv{More examples of Fashion++ edits}
     \item MTurk interfaces for the \KHarxiv{three} human subject studies provided in Section 4.2 of the main paper
    \item \KHarxiv{Full results and Turkers' verbal rationales (as a wordcloud) for user study A (Section 4.2 of the main paper)}
    \item Examples of Turkers' verbal descriptions of what actions to perform in user study C (Section 4.2 of the main paper)
\end{itemize}

\begin{figure*}
   \begin{center}
       \includegraphics[width=.32\linewidth]{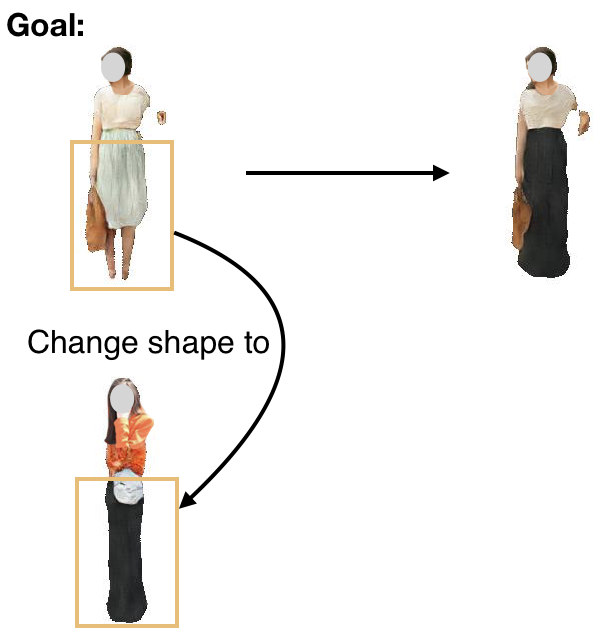}
       \includegraphics[width=.56\linewidth]{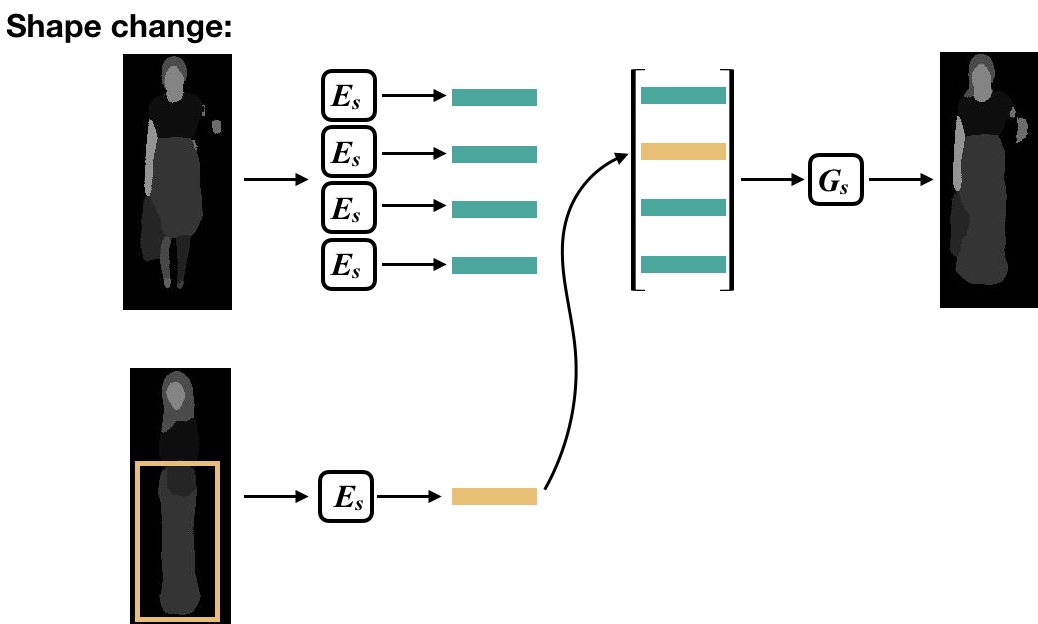}
    \end{center}
    \vspace*{-5mm}
    \caption{Shape generation using our shape-VAE. \KGsupp{In this example,} the goal is to change the girl's midi skirt to a long skirt. We encode each garment separately, overwrite the skirt's code with the code from the long skirt, and generate the final changed \KG{silhouette for the} outfit.}
   \label{fig:shape_generation}
    \vspace*{-4mm}
\end{figure*}

\section{Implementation details}
\paragraph{Training.}
We have two generators, a GAN for texture and a VAE for shape, and a classifier for editing operations.
All generation networks are trained from scratch, using the Adam solver~\cite{kingma2014adam} and a learning rate of $0.0002$.
For VAE, we keep the same learning rate for the first $100$ epochs and linearly decay the rate to zero over the next $200$ epochs.
For GAN, we keep the same learning rate for the first $100$ epochs and linearly decay the rate to zero over the next $100$ epochs.
For the fashionability classifier, we train from scratch with the Adam solver with weight decay $0.0001$ and a learning rate of $0.001$. We keep the same learning rate for the first $60$ epochs and decay it $10$ times every $20$ epochs until epoch $120$.

For the GAN, we adopt the architecture from~\cite{pix2pixHD2018}.
For the VAE,
our architecture is defined as follows:
Let $\mathtt{c7s1\mhyphen k}$ denote a $7\times 7$ convolutional block with $k$ filters and stride $1$. $\mathtt{dk}$ denotes a $3\times 3$ convolutional block with $k$ filters and stride $2$. $\mathtt{Rk}$ denotes a residual block that contains two $3\times 3$ convolutional blocks with $k$ filters. $\mathtt{pk}$ denotes a layer reflection padding $3$ on all boundaries. $\mathtt{fck}$ denotes a fully connected layer with $k$ filters. We use Instance Normalization (IN)~\cite{ulyanov2017IN} and ReLU activations.  The VAE consists of:
\begin{itemize}
\item Encoder: $\mathtt{p3}, \mathtt{c7s1}\mhyphen \mathtt{16}, \mathtt{d32}, \mathtt{d64}, \mathtt{d128}, \mathtt{d128}, \mathtt{d128},\\ \mathtt{d128}, \mathtt{d128}, \mathtt{R128}, \mathtt{R128}, \mathtt{R128}, \mathtt{R128}, \mathtt{R128}, \mathtt{R128}, \mathtt{R128},\\ \mathtt{R128}, \mathtt{R128}, \mathtt{fc8}$
\item Decoder: $\mathtt{d128}, \mathtt{d128}, \mathtt{d128}, \mathtt{d128}, \mathtt{d128}, \mathtt{d64}, \mathtt{d32}, \mathtt{d16},\\ \mathtt{p3}, \mathtt{c7s1\mhyphen 18}$
\end{itemize}
\KH{where the encoder is adapted from ~\cite{pix2pixHD2018} and decoder from ~\cite{BicycleGAN2017}.}
Our MLP for the fashionability classifier is defined as:\\
$\mathtt{fc256}, \mathtt{fc256}, \mathtt{fc128}, \mathtt{fc2}.$
For shape and texture features, both $d_s$ and $d_t$ are $8$. 
For the fashionability classifier to perform edits, we use an SGD solver with step size $0.1$. 

\paragraph{Baselines.}
Since the encodings' distribution of inventory garments is not necessarily Gaussian, the \textsc{random} baseline samples from inventory garments for automatic evaluation, and from a standard Gaussian for human subject study B.

\paragraph{Post-processing.}
As our system did not alter clothing-irrelevant regions, and to encourage viewers to focus on clothing itself, we automatically replace the generated hair/face region with the original, using their segmentation maps.

\section{Ablation \KG{study}}
We use $d_s, d_t=8$ throughout our paper. Here, we show the effect of texture and shape feature on their own, and how the dimension of the feature affects our model. We measure the feature's effect by the fashionability classifier (MLP)'s validation accuracy. 
We compare just using texture, just using shape, and using the concatenation of the two in \tabref{ablation}(a):
we found that shape is a more discriminative feature than texture.
We tried $d_t = 3,8$, and found that $d_t=8$ gives qualitatively more detailed images than $d_t=3$, but continuing increasing $d_t$ beyond $8$ does not give qualitatively better result. \tabref{ablation}(b) shows the feature dimension's effect \KG{on the quantitative results}, where left is just using the texture as the feature and right is concatenating both texture and shape feature. In both cases, increasing $d_t$ makes our features more discriminative.

\begin{table} 
   \centering
   \footnotesize
   \subfloat[\textbf{Feature selection.}\label{tab:ablation:feature_selection}]{
   \tablestyle{5pt}{1.2}\begin{tabular}{@{}ccc@{}}
      \multicolumn{2}{c}{} \\
      texture & shape & texture + shape \\
      \midrule
      0.663        &    0.741   & 0.751
   \end{tabular}}\hfill
   \subfloat[\textbf{Feature dimension.}\label{tab:ablation:feature_dimension}]{
   \tablestyle{5pt}{1.2}\begin{tabular}{@{\extracolsep{4pt}}cccc@{}}
      \multicolumn{2}{c}{texture} & \multicolumn{2}{c}{texture + shape} \\
      \cline{1-2}
      \cline{3-4}
      3         &        8             &             3         &           8 \\
      \midrule
      0.576     &       0.663          &    0.717              &       0.751
   \end{tabular}}
   \vspace*{-0.1in}
   \caption{Ablation study on how the outfit's \KG{features} affect the \KG{accuracy} of \KG{the} fashionability classifier.}
   \label{tab:ablation}
   \vspace{-4mm}
\end{table}

\section{More details about shape generation}
Here, we walk through the process of how our shape generator controls the silhouette of each garment. If our goal is to change an outfit's skirt, as in \figref{shape_generation} left, our shape encoder $E_s$ first encodes each garment separately, and then overwrites the skirt's code with the skirt we intend to change to. Finally, we concatenate each garment's code into $\ssb = [\ssb_0; \dots; \ssb_{n-1}]$, and our shape generator $G_s$ decodes it back to a region map. This process is shown in \figref{shape_generation} right.

\section{Automatic evaluation metric}
\begin{figure}
   \begin{center}
       \includegraphics[width=\linewidth]{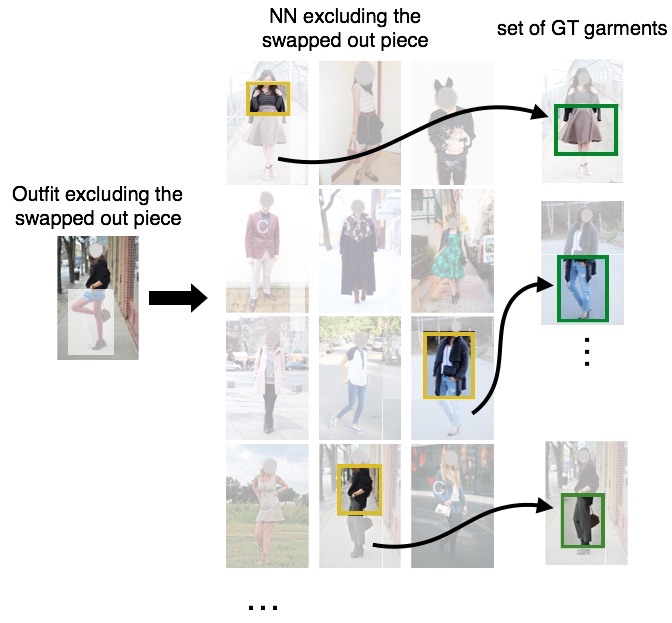}
    \end{center}
    \vspace*{-5mm}
    \caption{Formulating the set of GT garments per negative outfit.}
   \label{fig:gt_set}
\end{figure}

To automatically evaluate fashionability improvement, we need ground-truth (GT) garments to evaluate against. To capture multiple ways to improve an outfit, we form \emph{a set of GT garments} per outfit, as noted in Section 4.1 of the main paper.
Our insight is that the garments that go well with a given blouse appear in outfits that also have blouses similar to this one. As a result, we take the corresponding region's garments, that is the pants or skirts worn with these similar blouses, to form this set.
To do so, we first find the$M$ nearest neighbors of the unfashionable outfit \emph{excluding the swapped out piece} (\figref{gt_set} left), and then take the corresponding pieces in these $M$ neighbors (\figref{gt_set} right) as $M$ possible ways to make this outfit fashionable. We use the minimal distance of the original piece to all $K$ pieces in GT set as the original piece's distance to GT. Using median or mean gives similar results.

\begin{figure}
    \begin{center}
      \includegraphics[width=\linewidth]{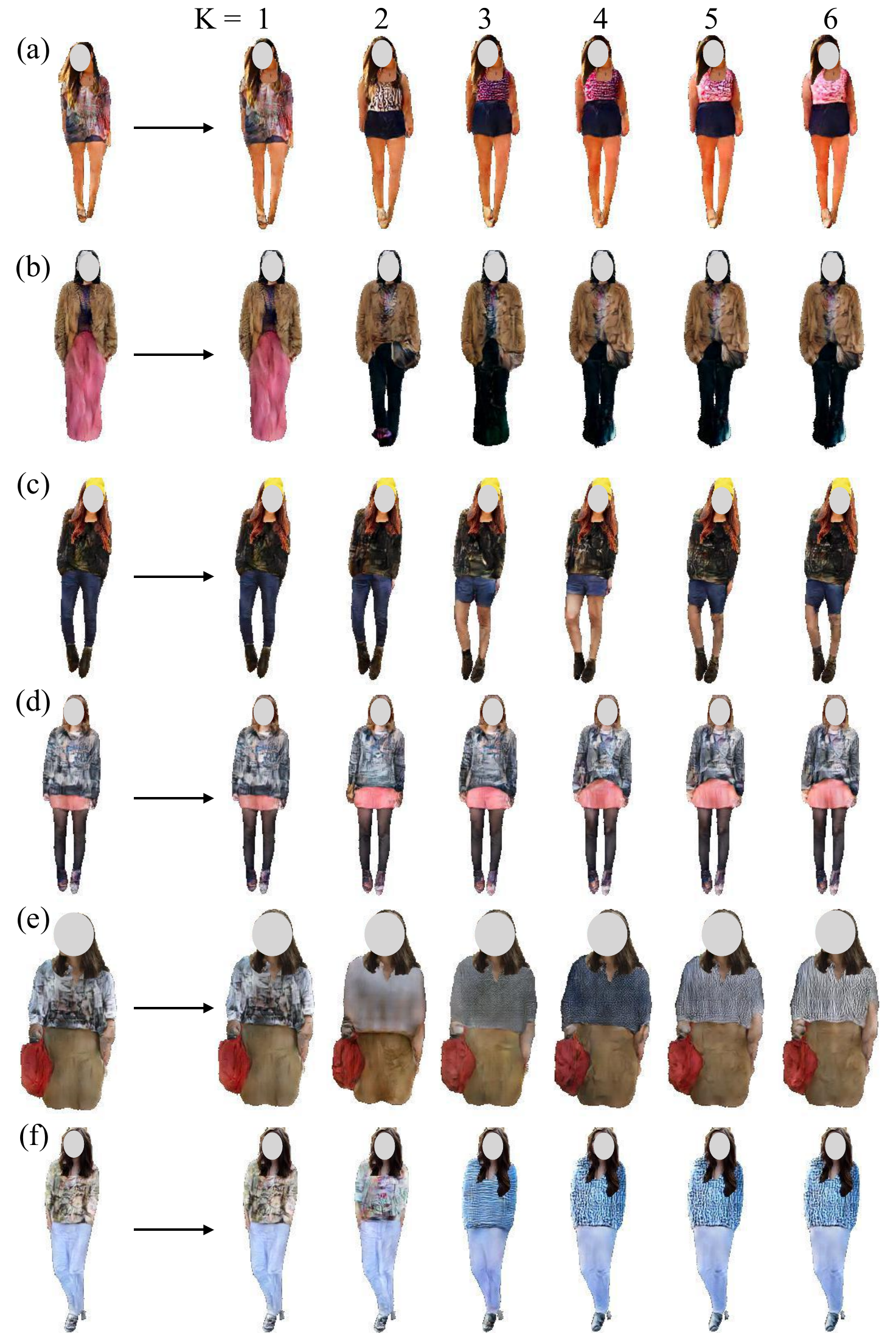}
    \end{center}
    \vspace*{-2mm}
    \caption{\KHarxiv{Spectrum of edits ($K=1$ to $6$) by Fashion++: the first column are the original outfits, and starting from the second are gradually editing the outfits more by taking more gradient steps, from $1$ to $6$.}}
   \label{fig:qual_ex_spectrum}
   \vspace*{-4mm}
\end{figure}

\section{More qualitative examples}
Due to the sake of space, we show one Fashion++ edit for each example in Section 4.3 of the main paper.
In \figref{qual_ex_spectrum}, we show more editing examples by Fashion++, and for each one we display the editing spectrum from $K=1$ to $6$.
\figref{qual_ex_spectrum}(a) is the full spectrum for one of the examples in Fig.~6 of the main paper. The outfit starts changing by becoming sleeveless and tucked in, and then colors become even brighter as more edits are allowed. (b) changes the pink long skirt to black flared pants, which actually are not too different in shape, but makes the outfit more energetic and better color matching. (c) gradually shortens the length of the jeans to shorts. (d) tucks in more amount of the sweater. Both (e) and (f) change the pattern of the blouses to match the bottom better. In most examples, edits typically start saturating after $K=4$, and changes are less obvious after $K=6$.

\section{Mechanical Turk Interface}
\figref{interface_A}, \figref{interface_B}, and \figref{interface_C} show our MTurk interfaces for the three human subject studies presented in the main paper. We  give them the definition of minimal editing and good/bad examples of edits, and tell them to ignore artifacts in synthesized images. 
For A, we ask them to (i) choose whether any of the changed outfits become more fashionable, and (ii) which is the best minimal edited outfit and (iii) why.
For B, we ask them two questions comparing the changed outfit to the original: (i) whether the changed outfit remains similar, and (ii) whether the changed outfit is more fashionable.
For C, we ask them if (i) they understand what to change given the original and changed outfit, and (ii) describe it verbally.

\begin{figure}
  \captionsetup[subfigure]{labelformat=empty}
  \footnotesize
  \subfloat[(a) \label{fig:user_study_A_fashion_improvement}] {
    \begin{tikzpicture}
      \centering
      \begin{axis}[
            ybar, axis on top,
            title={Fashion improvement},
            height=4cm, width=0.245\textwidth,
            bar width=0.2cm,
            ymajorgrids, tick align=inside,
            major grid style={draw=white},
            enlarge y limits={value=.1,upper},
            ymin=0, ymax=100,
            axis x line*=bottom,
            axis y line*=right,
            y axis line style={opacity=0},
            tickwidth=0pt,
            enlarge x limits=true,
            legend style={
                at={(0.5,-0.2)},
                anchor=north,
                legend columns=-1,
                /tikz/every even column/.append style={column sep=0.5cm}
            },
            ylabel={Percentage (\%)},
            xlabel={$K$},
            symbolic x coords={
               1,2,3,4,5,
               6,7,8,9,10}
              ]
        \addplot [draw=none, fill=blue!30] coordinates {
          (1,  1)
          (2,  20) 
          (3,  46)
          (4,  64) 
          (5,  68) 
          (6,  71)
          (7,  75) 
          (8,  81)
          (9,  85)
          (10, 92) };
      \end{axis}
    \end{tikzpicture}
  }\hfill
  \subfloat[(b) \label{fig:user_study_A_best_edit}] {
    \begin{tikzpicture}
      \centering
      \begin{axis}[
            ybar, axis on top,
            title={Best edit},
            height=4cm, width=0.245\textwidth,
            bar width=0.2cm,
            ymajorgrids, tick align=inside,
            major grid style={draw=white},
            enlarge y limits={value=.1,upper},
            ymin=0, ymax=25,
            axis x line*=bottom,
            axis y line*=right,
            y axis line style={opacity=0},
            tickwidth=0pt,
            enlarge x limits=true,
            legend style={
                at={(0.5,-0.2)},
                anchor=north,
                legend columns=-1,
                /tikz/every even column/.append style={column sep=0.5cm}
            },
            ylabel={Percentage (\%)},
            xlabel={$K$},
            symbolic x coords={
               1,2,3,4,5,
               6,7,8,9,10}
           ]
        \addplot [draw=none, fill=blue!30] coordinates {
          (1,  10.94890511)
          (2,  17.51824818) 
          (3,  14.72019465)
          (4,  11.67883212) 
          (5,  10.82725061) 
          (6,  14.11192214)
          (7,   5.596107056) 
          (8,   3.771289538)
          (9,   2.433090024)
          (10,  8.394160584) };
      \end{axis}
    \end{tikzpicture}
  }
  \caption{\KHarxiv{Results breaking down $K$ for user study A: (a) which of the changed outfits become more fashionable, and (b) which edited outfit makes the best minimal edit to the original outfit.}}
  \label{fig:user_study_A}
\end{figure}
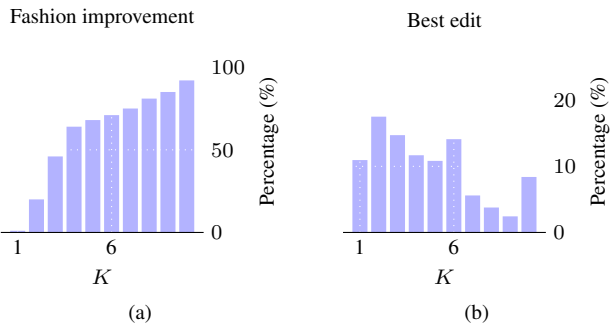

\section{Detailed result for user study A}
\KHarxiv{For question (i) \KGsupp{in user study A}, since there should be a consensus on fashionability improvement, we aggregate the responses over all subjects for each example. Each of the $100$ testing examples will be judged as either improved or not improved for every $K$.  The result is summarized in \figref{user_study_A_fashion_improvement}. As more changes are made (increasing $K$), more examples are rated as improving fashionability, with $92\%$ of them improved when $K=10$.}

\KHarxiv{Question (ii) is subjective in nature: different people prefer a different trade-off \KGsupp{(between the amount of change versus the amount of fashionability added)}, so we treat response from each subject individually.  The result is summarized in \figref{user_study_A_best_edit}. No specific $K$ dominates, and a tendency of preferring $K \leq 6$ is observed, in total $80\%$ of the time.}

\begin{figure}
    \begin{center}
      \includegraphics[width=\linewidth]{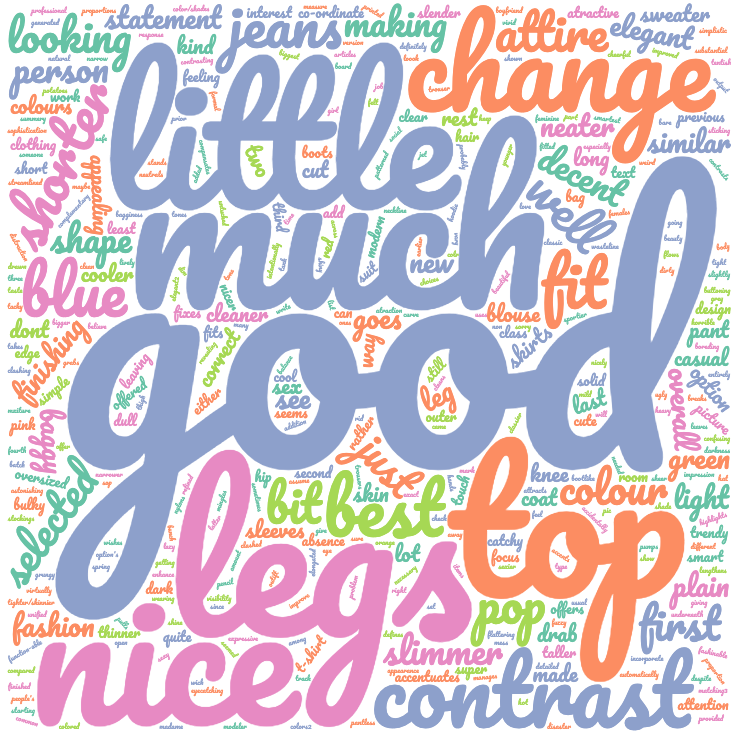}
     \end{center}
    \vspace*{-2mm}
    \caption{\KHarxiv{Summary in word cloud of why a changed outfit is preferred in user study A.}}
   \label{fig:verbal_explanation_wordcloud}
   \vspace*{-4mm}
\end{figure}

\begin{figure}
   \includegraphics[width=.88\linewidth]{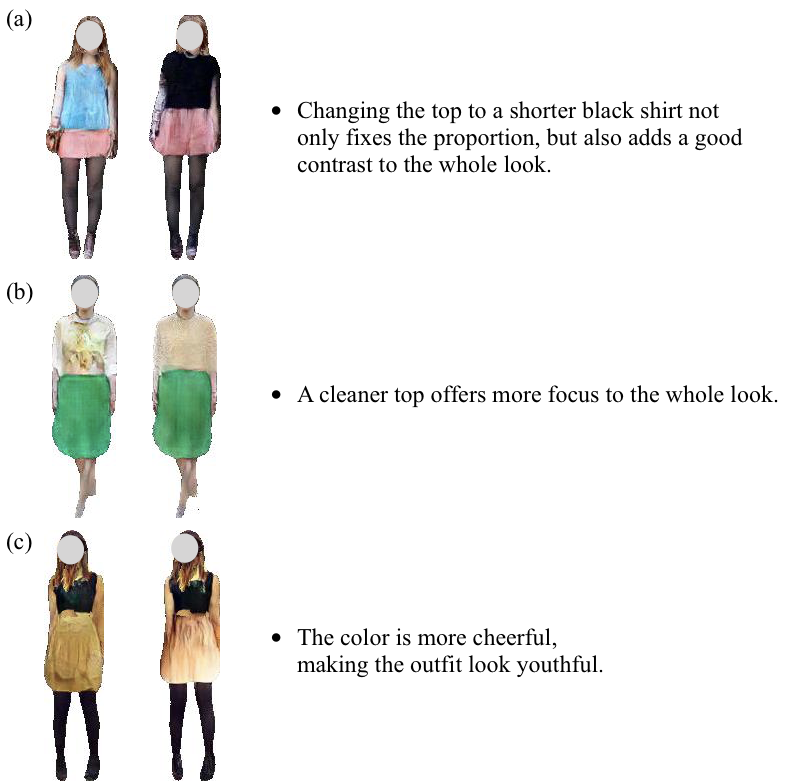}
       \includegraphics[width=\linewidth]{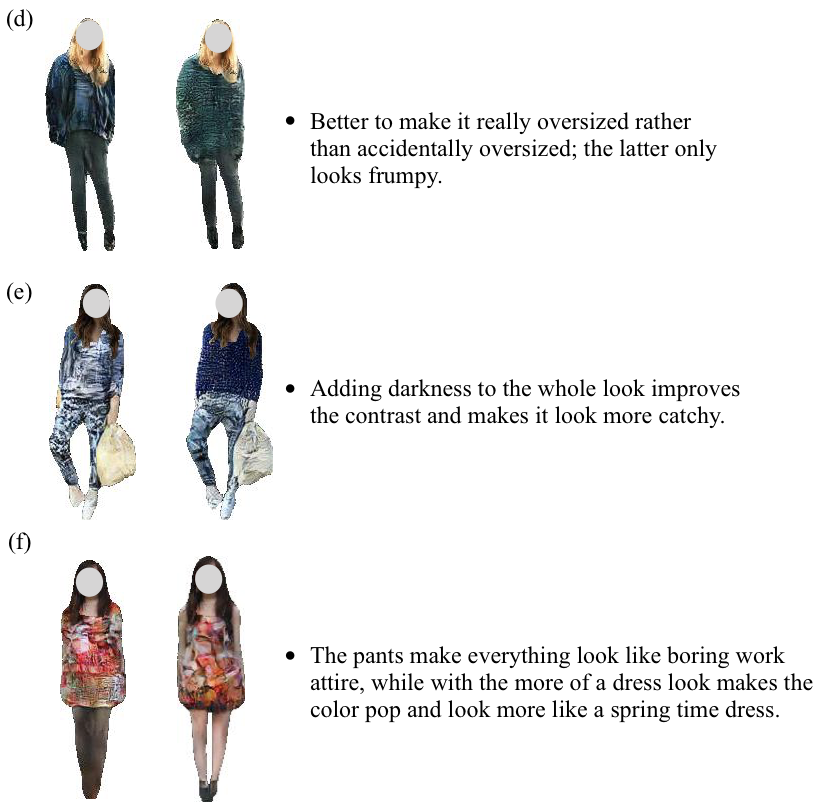}
    \vspace*{-2mm}
    \caption{\KGsupp{Examples of} \KHarxiv{Turkers' responses to user study A: pairs of images on the left show the original outfits and the changed outfits \KGsupp{generated by Fashion++} preferred by the user. Corresponding sentences on the right are their verbal explanation for why they make such selection.}}
   \label{fig:verbal_explanation_example}
    \vspace*{-4mm}
\end{figure}

\KHarxiv{For question (iii), we ask users their reasons to selecting a specific $K$ in question (ii). Examples of Turkers' responses are in \figref{verbal_explanation_example}. From phrases such as \emph{add contrast}, \emph{offer focus}, \emph{pop}, or \emph{catchy} in these examples, and a word cloud made from all responses (\figref{verbal_explanation_wordcloud}), we can tell that a common reason a user prefer an outfit is it being more attractive/interesting.}

\section{Verbal descriptions of actionable edits for user study C.}
\begin{figure}
       \includegraphics[width=.88\linewidth]{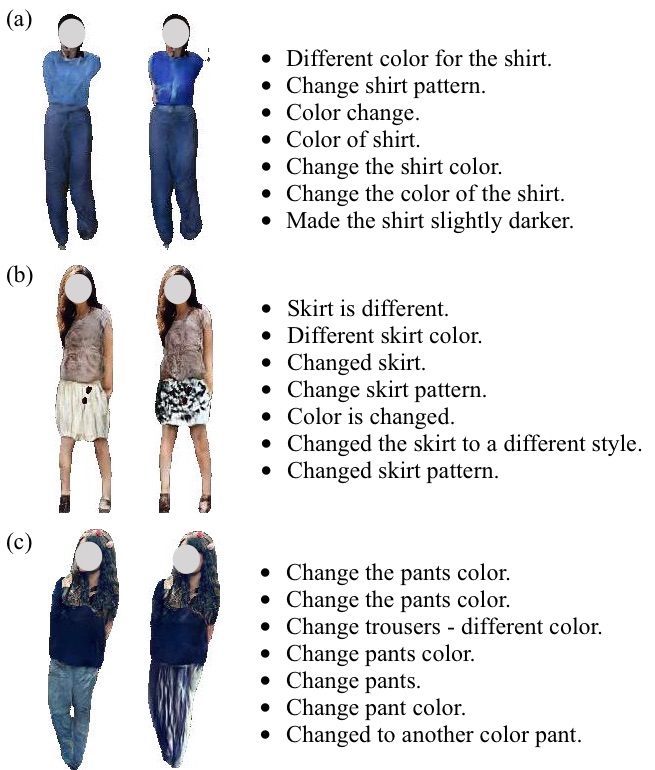}
       \includegraphics[width=\linewidth]{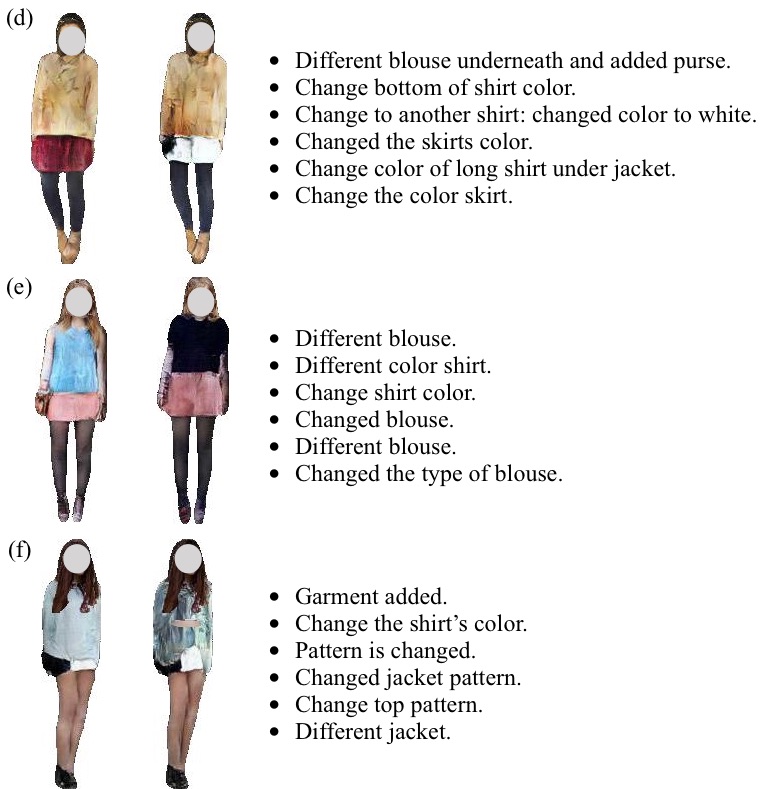}
       \vspace*{-2mm}
    \caption{Examples of Turkers' verbal descriptions about what is changed in a Fashion++ edit. Despite mild artifacts in the edits, note how humans reach consensus about what change is being recommended by the system.}
   \label{fig:verbal_description}
    \vspace*{-4mm}
\end{figure}

\begin{figure*}
   \begin{center}
       \includegraphics[width=.8\linewidth]{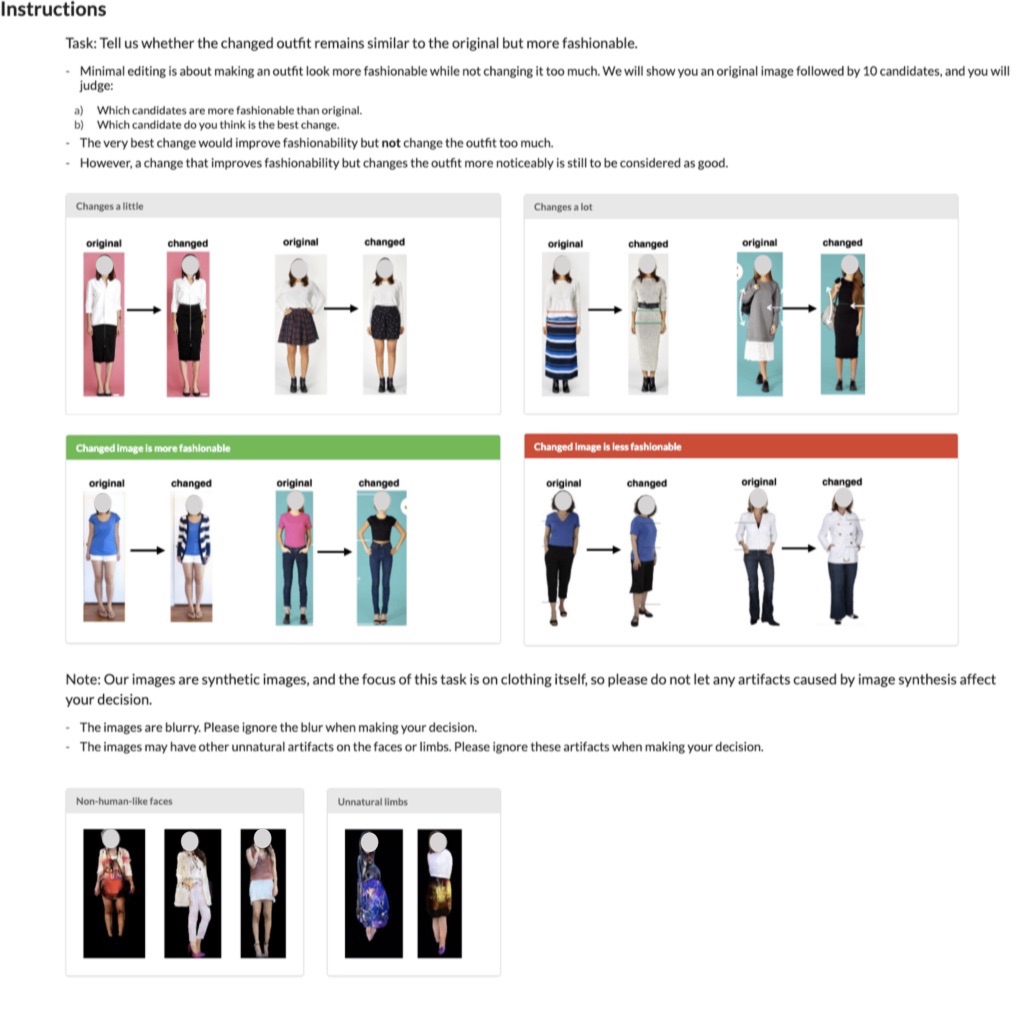}
       \includegraphics[width=.8\linewidth]{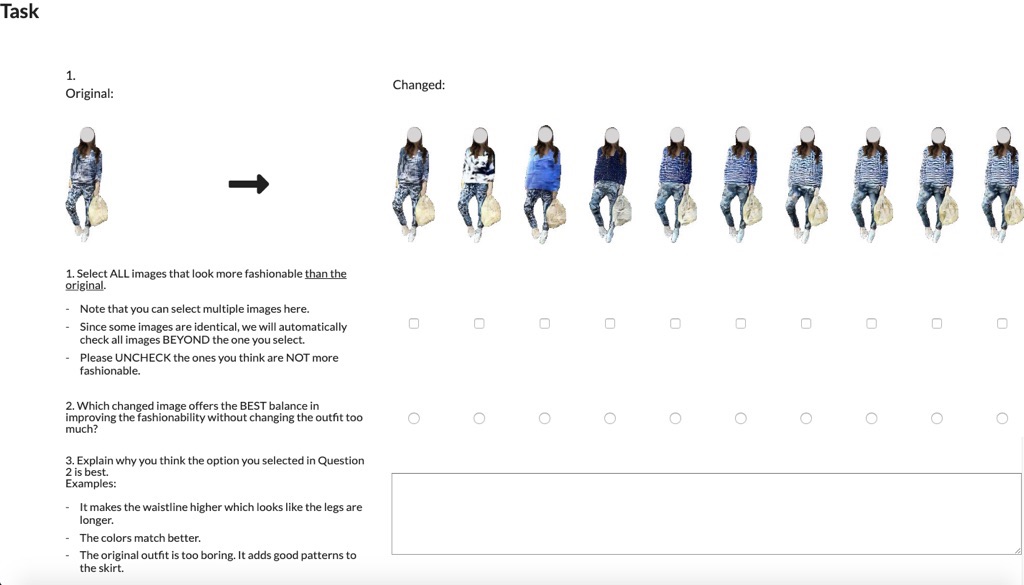}
    \end{center}
    \vspace*{-5mm}
    \caption{Interface for human subject study A: understanding to what extent a given degree of change is preferred and why.}
   \label{fig:interface_A}
\end{figure*}

\begin{figure*}
   \begin{center}
       \includegraphics[width=.9\linewidth]{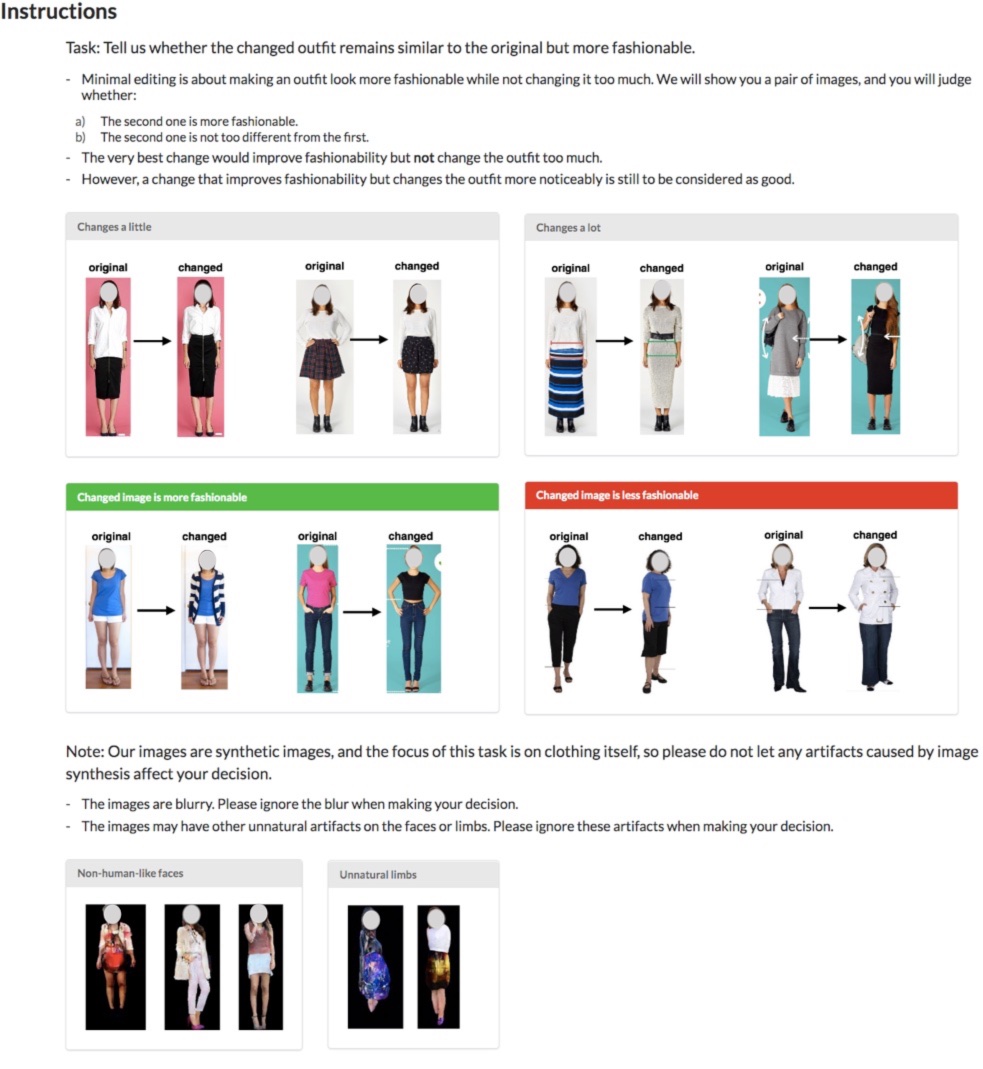}
       \includegraphics[width=.9\linewidth]{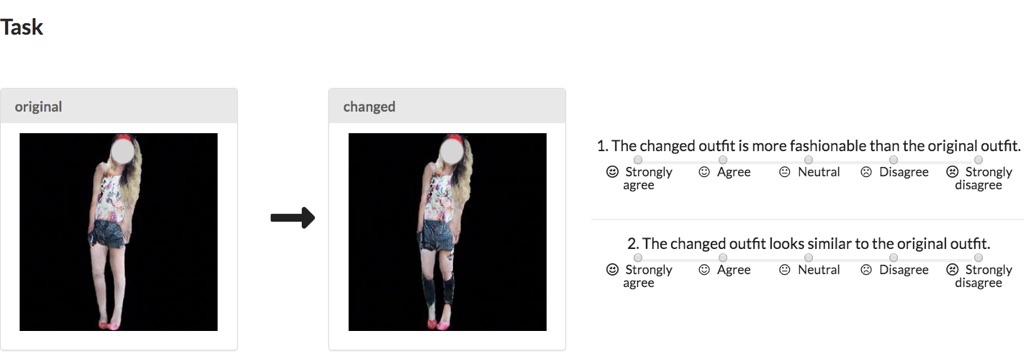}
    \end{center}
    \vspace*{-5mm}
    \caption{Interface for human subject study B: understanding how Fashion++ compares to the baselines.}
   \label{fig:interface_B}
\end{figure*}

\begin{figure*}
   \begin{center}
       \includegraphics[width=.9\linewidth]{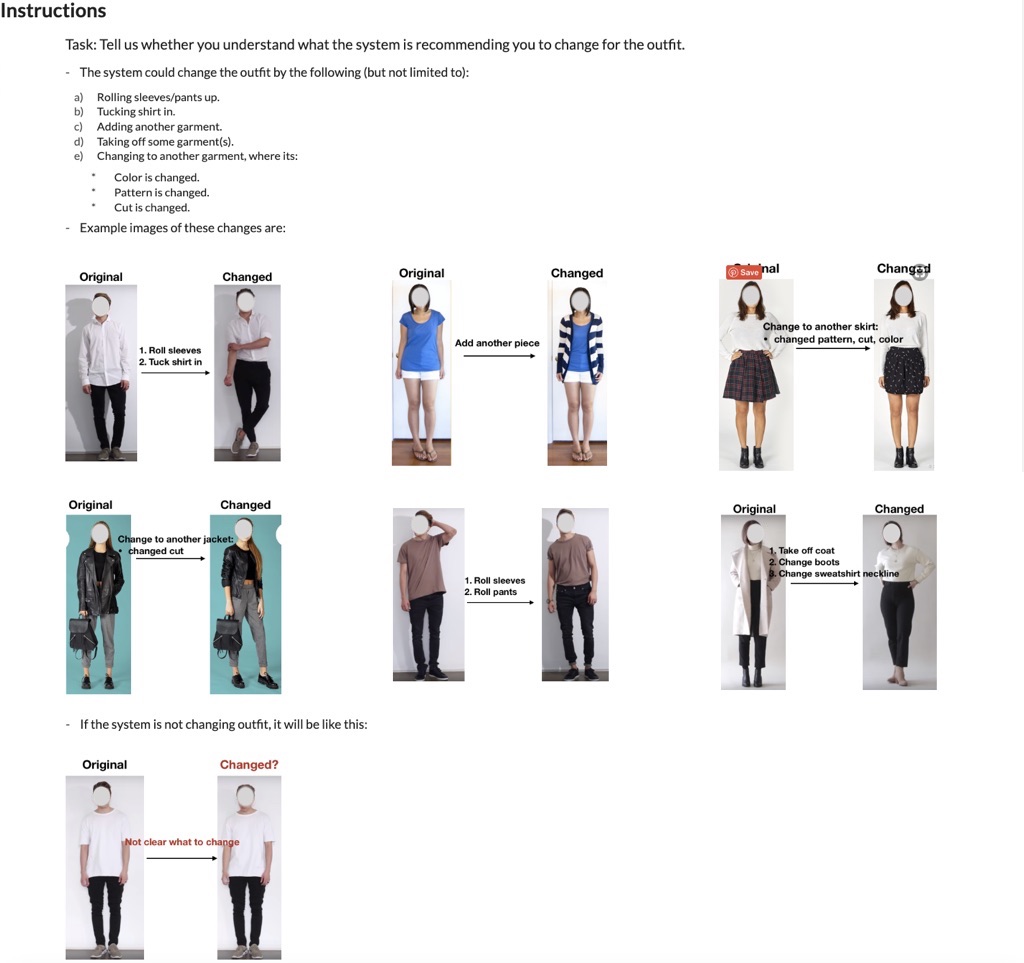}
       \includegraphics[width=.9\linewidth]{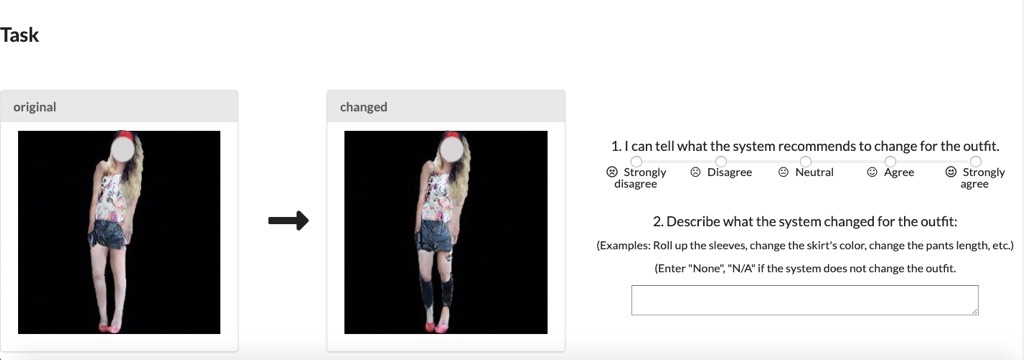}
    \end{center}
    \vspace*{-5mm}
    \caption{Interface for human subject study C: understanding whether humans can get \emph{actionable information} from the suggested edits.}
   \label{fig:interface_C}
\end{figure*}

In the experiment presented as user study C in the main paper, we asked Turkers to rate how actionable the suggested edit is, and briefly describe the edit in words.
\figref{verbal_description} shows example descriptions from human judges. Each example has 6 to 7 different descriptions from different people. 
For example, despite mild artifacts in \figref{verbal_description}(a), humans still reach consensus on the actionable information.  Note that in \figref{verbal_description}(b)(c)(d), most people described the edit as changing color/pattern, while in \figref{verbal_description}(e)(f) more descriptions are about changing to/adding another garment, because \figref{verbal_description}(e)(f) changes garments in a more drastic way.
\emph{Tweaking} the color/pattern of a garment is essentially changing to another garment, yet humans perceived this differently. When the overall style of the outfit remains similar, changing to a garment with different colors/patterns seems like a slight change to humans.

